%% file: root.tex
\documentclass[journal]{IEEEtran}

\usepackage{amsmath}
\usepackage{amssymb}
\usepackage{booktabs}
\usepackage[]{graphicx}
\usepackage[font=footnotesize]{caption}
\usepackage[font=footnotesize]{subcaption}
\usepackage{mathtools}
\usepackage{gensymb}
\usepackage{lipsum}
\usepackage{float}
\usepackage{color}
\usepackage{threeparttable}
\usepackage{todonotes}
\usepackage{mathtools}
\usepackage[]{algorithm2e}
\usepackage{soul}
\usepackage{siunitx}
\usepackage{hyperref}
\usepackage{multirow}
\usepackage{makecell}

\DeclareMathOperator*{\argmin}{arg\,min}

\newcommand{\Sec}{Section~}

\newcommand{\Fig}{Fig.~}
\newcommand{\Tab}{Table~}

\newcommand{\eg}{e.g., }
\newcommand{\ie}{i.e., }

\newcommand{\etal}{et al. }
\newcommand{\PAR}[1]{\vskip4pt \noindent{\bf #1~}}

\input{math_macros.tex}

\newcommand{\trace}{\text{Trace}}
\newcommand{\mineig}{\lambda_\text{min}}
\newcommand{\rrts}{\text{RRT}^{*}}
\newcommand{\fim}{\text{FIM}}
\newcommand{\vr}{\mathbf{v}^\text{r}}
\newcommand{\vp}{\mathbf{v}^\text{p}}
\newcommand{\z}{\mathbf{z}}

\newcommand{\poly}{\text{poly}}
\newcommand{\qd}{\text{quad}}
\newcommand{\gp}{\text{gp}}
\newcommand{\lin}{\text{lin}}

\begin{document}

\title{Fisher Information Field:\\an Efficient and Differentiable Map for Perception-aware Planning}

\author{Zichao Zhang, Davide Scaramuzza%
\thanks{The authors are with the Robotic and Perception Group, at both the
	Dep. of Informatics (University of Zurich) and the Dep. of Neuroinformatics
	(University of Zurich and ETH Zurich), Andreasstrasse 15, 8050 Zurich,
	Switzerland.}%
}

\maketitle

\begin{abstract}

Considering visual localization accuracy at the planning time gives preference to robot motion that can be better localized and thus has the potential of improving vision-based navigation, especially in visually degraded environments.
To integrate the knowledge about localization accuracy in motion planning algorithms, a central task is to quantify the amount of information that an image taken at a 6 degree-of-freedom pose brings for localization, which is often represented by the Fisher information.
However, computing the Fisher information from a set of sparse landmarks (\ie a point cloud),  which is the most common map for visual localization, is inefficient.
This approach scales linearly with the number of landmarks in the environment and does not allow the reuse of the computed Fisher information.
To overcome these drawbacks, we propose the first dedicated map representation for evaluating the Fisher information of 6 degree-of-freedom visual localization for perception-aware motion planning.
By formulating the Fisher information and sensor visibility carefully, we are able to separate the rotational invariant component from the Fisher information and store it in a voxel grid, namely the \textit{Fisher information field}.
This step only needs to be performed once for a known environment.
The Fisher information for arbitrary poses can then be computed from the field in \textit{constant} time, eliminating the need of costly iterating all the 3D landmarks at the planning time.
Experimental results show that the proposed Fisher information field can be applied to different motion planning algorithms and is at least one order-of-magnitude faster than using the point cloud directly.
Moreover, the proposed map representation is differentiable, resulting in better performance than the point cloud when used in trajectory optimization algorithms.
\end{abstract}


\IEEEpeerreviewmaketitle

\section*{supplementary material}
Video: \url{http://rpg.ifi.uzh.ch/fif.html}

Code: \url{https://github.com/uzh-rpg/rpg_information_field}

\input{sections/intro.tex}
\input{sections/related_work.tex}
\input{sections/preliminaries.tex}
\input{sections/standard_fisher_planning.tex}
\input{sections/approx_fim.tex}
\input{sections/fif_planning.tex}
\input{sections/experiments.tex}

\input{sections/conclusion.tex}

\bibliographystyle{IEEEtran}
\bibliography{rpg,references}   %

\end{document}

%% file: math_macros.tex
\newcommand{\R}{\mathtt{R}}
\newcommand{\expm}{\exp}

\global\long\def\se{\mathfrak{{se}}}
\global\long\def\SE{\text{{SE}}}

\newcommand{\secoords}{\pmb{\xi}}

\newcommand{\px}{\mathbf{u}}
\global\long\def\f{\mathbf{{f}}}
\global\long\def\pt{\mathbf{{p}}}
\global\long\def\proj{\operatorname{proj}}
\global\long\def\T{\mathtt{{T}}}

\global\long\def\I{\mathtt{{I}}}
\global\long\def\J{\mathtt{{J}}}
\global\long\def\H{\mathtt{{H}}}

\global\long\def\A{\mathtt{{A}}}

\global\long\def\C{\mathtt{{C}}}
\global\long\def\V{\mathtt{{V}}}

\global\long\def\Twc{\T_{\texttt{{wc}}}}
\global\long\def\Tcw{\T_{\texttt{{cw}}}}
\global\long\def\pc{\pt^{\texttt{{c}}}}
\global\long\def\pw{\pt^{\texttt{{w}}}}

\global\long\def\K{\mathtt{{K}}}

\global\long\def\prob{\text{p}}
\global\long\def\meas{\mathbf{z}}
\global\long\def\param{\mathbf{x}}

\global\long\def\bm{\mathbf{m}}

\global\long\def\twc{\mathbf{t}_{\texttt{wc}}}
\global\long\def\Rwc{\R_{\texttt{wc}}}
\global\long\def\Rcw{\R_{\texttt{cw}}}
\global\long\def\S{\mathtt{{S}}}
\DeclarePairedDelimiterX{\norm}[1]{\lVert}{\rVert}{#1}
\DeclarePairedDelimiterX{\skewsym}[1]{[}{]_{\times}}{#1}
\global\long\def\Id{\mathcal{{I}}}
\global\long\def\e{\mathbf{e}}
\DeclareMathOperator{\diag}{diag}
\DeclareMathOperator{\Tr}{Tr}

%% file: sections/intro.tex
\section{Introduction}\label{sec:intro}
One unique aspect of robot vision is that the robot has the ability to affect the data acquisition process by controlling the motion of the cameras.
Specifically, the robot motion impacts the information that will be captured by the cameras and thus influences the performance of perception algorithms.
Therefore, taking into consideration the requirement of perception algorithms in motion planning can potentially improve the performance of tasks that rely on visual perception - this is known as \emph{active perception} \cite{Bajcsy88ieee}.
Various tasks have been shown to benefit from active perception, such as state-estimation \cite{Costante16arxiv, Zhang18icra}, reconstruction \cite{Isler16icra}, exploration and navigation \cite{Papachristos17icra, Zhou20arxiv}, and manipulation \cite{zaky20arxiv}.
In this paper, we are particularly interested in the task of \textit{visual localization}, which estimates the 6 Degree-of-Freedom (DoF) camera pose from which a given image was taken relative to a reference scene representation.
We refer to the process of considering the quality of visual localization in motion planning as \emph{active visual localization}.
Active visual localization, or more generally active visual simultaneous localization and mapping (SLAM), is still an open problem \cite{Cadena16tro}.
From a practical perspective, when a robot operates in a (partially) known environment, visual localization is often used to bound the drift accumulated in local motion estimation such as visual-inertial odometry (VIO) \cite{MurArtal17tro, Qin18tro}.
Being able to plan the robot motion so that the robot can be better localized with respect to, \eg known landmarks, will be beneficial to maintain a reliable state estimation.

\begin{figure}[t]
	\centering
	\includegraphics[width=0.9\linewidth]{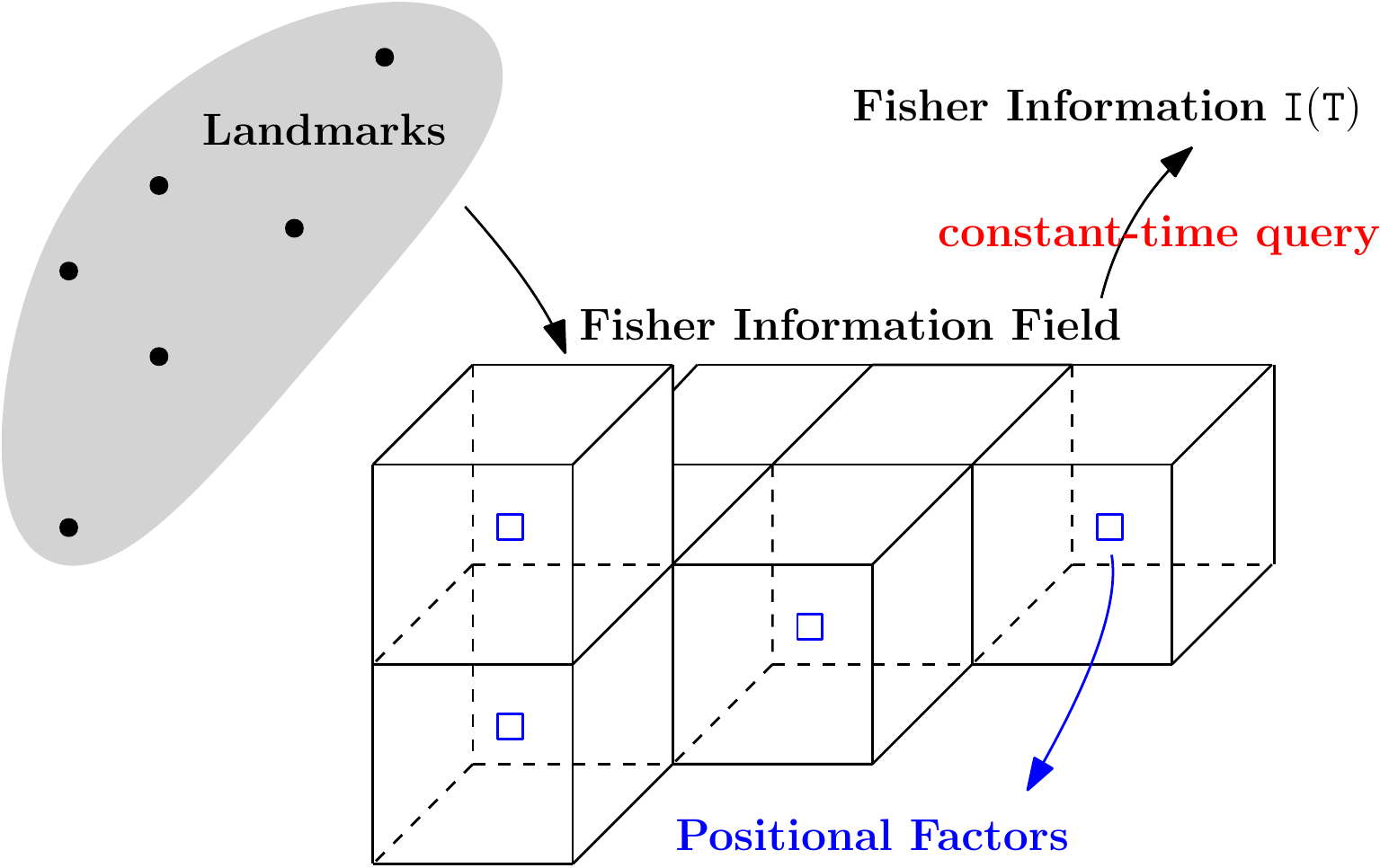}
	\caption{
		Illustration of the proposed Fisher information field.
		The gray cloud denotes the 3D landmarks in the environment.
		For each voxel (black cubes), the building process summarizes the positional (rotation-independent) information factors \eqref{eq:general_info_pif} or \eqref{eq:general_trace_pif} (blue squares).
		Then the information of an arbitrary pose $\T$ can be computed in \emph{constant time} without accessing the original 3D landmarks.
	}
	\label{fig:vox_mapping}
\end{figure}

One major paradigm in the literature for active visual localization is to plan the sensor motion based on the Fisher information/covariance in the corresponding estimation problem \cite{Indelman15ijrr, Costante16arxiv, Alzugaray17icra, Zhang18icra}.
Specifically, how well a certain motion (\eg a trajectory) can be localized in a known map is represented by the localizaiton quality of the individual poses sampled from the motion and incorporated in motion planning algorithms, for example, as part of the objective function to maximize.
The ``localizability'' of the poses (\ie how well/uncertain the pose can be estimated) is quantified by the Fisher information of the pose estimation problem.
For a landmark-based map, which is arguably the most common scene representation for visual localization, the 6 DoF pose is typically estimated from the observations of the landmarks in the image, and thus the corresponding Fisher information needs to be computed by iterating over all the landmarks in the map to account for the contribution of each landmark.
This approach, in spite of the convenience of using the same scene representation as visual localization (\ie point clouds), exhibits several limitations.
\textbf{First}, to evaluate the localizability of a single pose, one needs to evaluate the information for \emph{all} the 3D points, the complexity of which increases \textit{linearly} with the number of landmarks.
\textbf{Second}, this process has to be repeated many times in both sampling-based (\ie evaluating motion samples) and optimization-based methods (\ie optimization iterations), which introduces redundant computation, especially when the planning is performed multiple times in the same environment.
\textbf{Third}, due to the discontinuity of the actual visibility (see \Fig\ref{fig:vis_approx}), the Fisher information and related metrics are not differentiable with respect to the 6 DoF pose, which is not ideal for optimization-based motion planning algorithms.
These limitations indicate that point clouds, as a natural representation in SLAM/localization, is not ideal for the task of active visual localization.
Unfortunately, there is little work in designing dedicated scene representations for computing the Fisher information of 6 DoF  localization.

In view of the aforementioned limitations of point clouds, we propose a dedicated scene representation, namely \emph{Fisher Information Field} (FIF), for querying the Fisher information of arbitrary 6 DoF poses.
Specifically, the scene is represented as a voxel grid.
For each voxel, we summarize a \textit{rotation-independent} component of the Fisher information from all the 3D landmarks and store it in the voxel, which is applicable for all the poses that fall in this voxel, regardless of the orientation.
At query (\eg motion planning) time, given a 6 DoF pose, we first get the corresponding voxel via voxel hashing \cite{NieBner13tog}, and then the full Fisher information (under some approximation) of this pose can be recovered by applying a linear transformation to the stored rotation-independent component. The computing of the Fisher information for a pose is thus of \textit{constant} time complexity instead of linear.
Moreover, since the proposed FIF is precomputed in a voxel grid, it can be used for multiple planning sessions.
It can also be easily updated when landmarks are added to or deleted from the environment.

The idea of using a voxel grid is similar to Euclidean Signed Distance Field (ESDF) \cite{Oleynikova17iros} for collision avoidance, which stores in each voxel the distance to the closest point.
However, the key difference/difficulty is that the Fisher information additionally depends on the camera orientation due to the fact that the visibility of landmarks can vary drastically with orientations.
We therefore propose a novel formulation of the Fisher information that allows the aforementioned separation and pre-computation, which is key to the efficient query process.
The formulation is also differentiable, making our map representation suitable for optimization-based motion planning algorithms.

In summary, the proposed FIF overcomes the aforementioned limitations of using point clouds to compute the Fisher information of 6 DoF poses.
To the best of our knowledge, this is the first dedicated map representation that is capable of computing the Fisher information of 6 DoF localization efficiently.
Our map representation is general and can be integrated with different motion planning algorithms.
Experiments with both sampling-based and optimization-based methods demonstrate that FIF is up to two order of magnitude faster than point clouds in a typical motion planning scenario.
The performance, in terms of the localization success rate and accuracy of the planned motion, is comparable to point clouds.
Furthermore, when used in trajectory optimization,
the proposed map representation, in addition to being far more efficient, achieves better localization accuracy, due to its differentiability.

The core contribution of this work is thus a general map representation for perception-aware planning, which is differentiable and far more efficient at planning time than the standard practice of using point clouds.
This paper is an extension of our previous paper \cite{Zhang19icra}. The novelty of the present work includes:
\begin{itemize}
	\item A non-parametric visibility approximation that is more accurate and scalable than the quadratic function in the previous work.
	\item A novel way of computing the thresholds and potential costs related to Fisher information for perception-aware planning.
	\item Demonstration of using the proposed FIF in both sampling-based and optimization-based motion planning algorithms.
	\item Extension of the open source code to include the novel contributions.
\end{itemize}

The rest of the paper is structured as follows.
After reviewing the related work in \Sec\ref{sec:related_work}, we briefly introduce the Fisher information matrix and Gaussian process regression in \Sec\ref{sec:preliminiaries} as preliminaries for our approach.
In \Sec\ref{sec:fim_plan_standard}, we describe how Fisher information is typically used in a perception-aware planning setup and hightlight the limitations of computing the Fisher information from point clouds directly.
Then we introduce our formulation of the Fisher information in \Sec\ref{sec:fim_approx} and how it can be used to design a dedicated map representation for motion planning in \Sec\ref{sec:fif_planning}.
In \Sec\ref{sec:exp}, we present detailed experimental results regarding the properties of the proposed Fisher information field and its application to different motion planning algorithms.
Finally, we conclude the paper with some discussion about our method and possible future directions in \Sec\ref{sec:conclusion}.

%% file: sections/related_work.tex
\section{Related Work}
\label{sec:related_work}
\subsection{Perception-aware Motion Planning}
Considering perception performance in planning has been extensively studied in different contexts.
Early works include
maximizing the Fisher information (or equally minimizing the covariance) about the robot state and the map in navigation tasks \cite{Feder99ijrr, Alexei02iros},
minimizing the entropy of the robot state in known environments \cite{Burgard97ijcai, Roy99icra}, and actively searching features in SLAM systems \cite{Davison02pami}.
Recently, with the advance of drones, several works have been done to couple perception, planning and control on agile aerial platforms \cite{Achtelik13icra, Sadat14icra, Mostegel14icra, Papachristos17icra, Alzugaray17icra, Zhang18icra, Watterson18rss, Falanga18iros}.

Despite the extreme diversity of the research in this topic, related work can be categorized based on the method to generate motion profiles.
One paradigm uses sampling-based methods, which discretize the space of possible motions and find the optimal one in a discrete set.
Roy \etal \cite{Roy99icra} used the Dijkstra's algorithm to find the path on a grid that minimizes a combined cost of collision and localization.
Papachristos \etal \cite{Papachristos17icra} and Costante \etal \cite{Costante16arxiv} adapted the rapidly-exploring random tree (RRT) algorithms to incorporate the perception cost, and the latter additionally considered the photometric property of the environment.
Alzugaray \etal \cite{Alzugaray17icra} sampled positions near obstacles based on the intuition that pose estimation
error is small when the camera is close to the features on obstacles. Then path planning was carried out based on the sampled positions.
Zhang \etal \cite{Zhang18icra} proposed to evaluate motion primitives against multiple costs, including the localization uncertainty, in a receding horizon fashion.
Instead of a combined cost, as in most of previous works, Ichter \etal \cite{Ichter17isrr} used multi-objective search for perception-aware planning.

Alternatively, researchers have explored to plan in the continuous motion space.
Indelman \etal \cite{Indelman15ijrr} considered optimizing the motion within a finite horizon to minimize a joint cost including the final pose covariance, which was later extended to visual-inertial sensing and self-calibration in \cite{Elisha17iros}.
Watterson \etal \cite{Watterson18rss} studied the general problem of trajectory optimization on manifolds and applied their method to planning under the field-of-view (FoV) constraint of the camera.
The perception constraint can also be used at the controller level.
Falanga \etal \cite{Falanga18iros} integrated the objective of maximizing the visibility of a point of interest and minimizing its velocity in the image plane as the costs in model predictive control (MPC).
Lee \etal \cite{Lee20ral} trained a neural network to predict the dynamics of the pixels on the objects of interest (\eg gates in drone racing) and incorporated such information in a MPC framework.
Greeff \etal \cite{Greeff20ifac} considered the perception task of visual localization in a teach-and-repeat setup.
They modeled the probability of whether a landmark can be matched considering the perspective change and FoV constraint, and used the model in a MPC controller.
In the context of drone racing, there is also work that considers the time optimality in trajectory generation or optimal control, in addition to perception constraints.
Murali \etal \cite{Murali19acc} generated the position trajectory by considering collision constraints, and optimized the yaw considering the co-visibility of certain landmarks and the execution time of the trajectory.
Spasojevi \etal \cite{Spasojevic20icra} further proposed a trajectory parameterization algorithm that considers the FoV constraints and optimizes the traverse time at the same time.

In the aforementioned work, the perception related cost/metric were always calculated from a sparse set of 3D points in the environment.
As noted in Alexei \etal \cite{Alexei02iros}, calculating the metric (\eg ``localizability'') is an expensive operation, which we believe is due to the lack of proper map representations.
Unfortunately, little work has been done in developing dedicated representations for the efficient computation of related metrics, which is the primary contribution of this work.
Next, we further review some related work in map representations for perception quality and other related tasks.

\subsection{Map Representations}
Roy \etal \cite{Roy99icra} pre-computed and stored the information in a 2D grid, but their method was limited to $360^\circ$ FoV sensors.
Specifically, the visual information (\eg visibility) are invariant regardless of the camera orientation for omnidirectional sensors, and thus their map did not need to consider the impact of orientations, which is not true for cameras with limited FoVs.
More recently, Ichter \etal \cite{Ichter17isrr} trained a neural network to predict the state estimation error and generated a map of perception cost using the network prediction.
However, their map only contains the averaged cost of different orientations and, therefore, cannot be used to evaluate the cost of an arbitrary 6 DoF pose.
In contrast, our method explicitly models the FoV constraint and can represent the information of 6 DoF poses efficiently.
As a concurrent work, Fey \etal \cite{Fey19arxiv} proposed the similar idea of combining the information from many landmarks for efficient online inference in the context of trajectory optimization.
In contrast, our work focuses on a general map representation that is applicable to different motion planning algorithms.

Our approach is also connected to a couple of map representations for other tasks.
It is partially inspired by the approach of using ESDF for collision-free motion planning \cite{Oleynikova16rssw}.
Conceptually, both ESDF and our method summarizes the information from many 3D points/landmarks into a compact field (in the form of a voxel grid) for efficient query.
In the context of computer graphics, a common technique to speedup the rendering process is \emph{precomputed visibility volume} \cite{ue_pvv}.
Basically, the scene is first divided into cells, and for each cell, the visibility states of the static objects from this cell are precomputed and stored before the rendering process.
Then at rendering time, whether to render a specific object can be efficiently determined from the precomputed values.
The precomputed visibility volume reduces the rendering time at the cost of increasing runtime memory.
It is conceptually similar to our approach, where we achieve efficient query of the Fisher information matrix at the cost of more memory usage.

%% file: sections/preliminaries.tex
\section{Preliminaries}
\label{sec:preliminiaries}
\subsection{The Fisher Information Matrix}
For a general parameter estimation problem, the Fisher information matrix (FIM) summarizes the information that the observations carry about the parameters to be estimated.
To put it formally, if the measurement process can be described as a conditional probability density function
$\prob(\meas | \param)$, where $\meas$ is the measurement and $\param$ the parameters, one definition\footnote{The presented definition is the \textit{observed Fisher information}. See \cite{Efron1978biometrika} for the comparison of different concepts.} of the Fisher information is
\begin{eqnarray}
\I_{\param}(\meas) = (\frac{\partial}{\partial\param}\log\prob(\meas|\param))^{\top}
(\frac{\partial}{\partial\param}\log\prob(\meas|\param)).\label{eq:fim}
\end{eqnarray}
With identical and independent zero-mean Gaussian noise $\mathcal{N}(0, \sigma^2)$ on the measurement, \eqref{eq:fim} can be written as 
\begin{eqnarray}
\I_{\param}(\meas) = \frac{1}{\sigma^2}(\J_\param)^{\top}\J_\param,\quad
\text{where} \; \J_\param = \frac{\partial\meas}{\partial\param}.\label{eq:fim_gaussian}
\end{eqnarray}
Note that in practice \eqref{eq:fim} and \eqref{eq:fim_gaussian} are usually evaluated at the estimate $\param^{*}$ instead of the unknown true value $\param$.

The Fisher information is a pivotal concept in parameter estimation problems.
Most notably, the inverse of the FIM defines the Cram{\'e}r-Rao lower bound, which is the smallest covariance (in terms of Loewner order) that can be achieved by an unbiased estimator \cite[App.~3.2]{Hartley03book} \cite[p.~14]{Barfoot17book}.
Note that the widely used nonlinear maximum likelihood estimator (MLE) is in general biased, but the bias also tends to decrease when the Fisher information increases \cite{Box1971jrss}.
Due to its rich theoretical implications, the FIM is widely used in different applications, such as optimal design of experiments \cite{Friedrich06book}, active SLAM \cite{Indelman15ijrr} and various algorithms for information selection in perception systems \cite{Carlone17icra, Hsiung18iros, Usenko20ral, Kuo20icra}.

\subsection{Gaussian Process Regression}
\label{subsec:pre_gp_reg}
A Gaussian process (GP) is a collection of random variables, and any subset of them has a joint Gaussian distribution \cite{Rasmussen05gpml}.
In the context of a regression task, suppose we know the samples at 
$\mathbf{z} = \{z_i\}_{i=1}^{P}$ with the output $\mathbf{y} = \{y_i\}_{i=1}^{P}$,
and we would like to know the output value $y^*$ at $z^*$.
Under the assumption of Gaussian process, we have
\begin{equation}
\begin{bmatrix}
\mathbf{y} \\
y^*
\end{bmatrix}
\sim
\mathcal{N}(\mathbf{0},
\begin{bmatrix}
\K_{\mathbf{z}\mathbf{z}}& \K_{\mathbf{z}z^*}\\
\K_{{z^*}\mathbf{z}}& k(z^*, z^*)
\end{bmatrix}
),
\label{eq:gp_prior}
\end{equation}
where $\K_{\z\z}$ is of $P \times P$, $\K_{\z z^{*}}$ is of $P \times 1$ and $\K_{z^{*}\z} = \K^{\top}_{\z z^{*}}$.
In particular, $\K_{\mathbf{z}\mathbf{z}}^{i, j} = k(z_i, z_j)$ and 
$\K_{z^*\z}^{i} = k(z^*, z_i)$.
Then the GP regression simply takes the conditional distribution
\begin{equation}
y^* \sim \mathcal{N}(
\K_{{z^*}\mathbf{z}} \K_{\mathbf{z}\mathbf{z}}^{-1} \mathbf{y},
k(z^*, z^*) -
\K_{{z^*}\mathbf{z}} \K_{\mathbf{z}\mathbf{z}}^{-1} \K_{\mathbf{z}{z^*}}),
\label{eq:gp_regress}
\end{equation}
which gives both the regressed value and variance.

Obviously, the properties of the prior \eqref{eq:gp_prior} and the regressed result \eqref{eq:gp_regress} depends on the function $k(\cdot)$.
$k(a, b)$ is called the \textit{kernel function}, and intuitively encodes the correlation of the outputs at $a, b$.
Often $k(\cdot)$ is a parameterized functions, whose parameters are the \textit{hyperparameters} of a GP.
Perhaps the most used kernel function is the \emph{Squared Exponential kernel}:
\begin{equation}
k_{\text{SE}}(a, b) =  \sigma^2 \exp(-\frac{(a-b)^2}{2l^2}),
\end{equation}
where $\sigma$ and $l$ are the hyperparameters and can be calculated by maximizing the likelihood of the training data $\mathbf{z}$ and $\mathbf{y}$.

For simplicity, the above introduction is limited to the case where both the output and input are scalars. However, GP can also be generalized to vector input and output.
For a thorough description of GP (\eg optimization of hyperparameters), we refer the reader to \cite{Rasmussen05gpml}.
GP is a flexible model that finds many applications in robotics (\eg motion planning \cite{Dong16rss}, state estimation \cite{Barfoot14rss}).
In this work, we use GP to approximate the visibility of a landmark observed from a camera located at a certain pose, as detailed in \Sec\ref{subsec:vis_approx}.

%% file: sections/standard_fisher_planning.tex
\section{Planning with FIM: Standard approach}
\label{sec:fim_plan_standard}

To take localization quality into consideration, a common practice is to incorporate the FIM in the motion planning algorithm.
Without the loss of generality, we denote the motion as a continuous time function $f(t;\bm)$, parameterized with $\bm$. The output of the function is the 6 DoF pose of the camera at a given time.
We can then formulate a perception-aware motion planning algorithm as:
\begin{eqnarray}
\bm^{*} = \underset{\bm}{\argmin}\; \mu_{\text{v}}C_{\text{v}}(f(t;\bm)) +
\mu_{\text{o}}C_{\text{o}}(f(t;\bm)),
\label{eq:active_loc_full}\vspace{-0ex}
\end{eqnarray}
where $C_{\text{v}}$ is the cost related to visual localization, $C_{\text{o}}$ denotes the other cost terms collectively (\eg collision and execution time) and $\mu_\text{v}$/$\mu_\text{o}$ are the corresponding weights.
Since localization can be viewed as the estimation of the poses of interest, the FIM can be used to quantify the estimation error and, thus, the localization quality.
Evaluating the cost using $M$ discrete samples, we have
\begin{eqnarray}
C_{\text{v}} = -s(
\begin{bmatrix}
\I_{\T_{1}}&0&0\\
0&\ldots&0\\
0&0&\I_{\T_{M}}\\
\end{bmatrix}),\quad
\I_{\T_i} = \sum_{k}^{k \in V_i}\I_{\T_{i}}(\pw_{k})
\label{eq:cost_info}
\end{eqnarray}
where $\T_i$ is the $i$th sample pose, $V_i$ is the index set of the landmarks that can be matched in the image taken at $\T_i$ (can be determined using, \eg \cite{Alcantarilla10icra,Alcantarilla11icra}), and $\pw_{k}$ is the $k$th landmark in the world frame.
$\I_{\T_i}(\pw_k)$ is calculated as in \eqref{eq:fim_gaussian} using the Jacobian of the observation of the $k$th landmark with respect to $\T_i$ (we limit our discussion to 3D structure-based localization).
$s(\cdot)$ is a metric function that converts the information matrix into a scalar.
It can be different metrics from the FIM directly, such as the determinant and the smallest eigenvalue of the FIM, which we refer to as \emph{information metrics}.
It is also possible to design and use other types of functions as $s(\cdot)$ (\eg see \Sec\ref{subsec:fif_motion_planning}).

\eqref{eq:active_loc_full} can be solved using sampling-based methods,
such as RRT \cite{Papachristos17icra} and motion primitives \cite{Zhang18icra}, or optimization-based methods \cite{Indelman15ijrr}.
Either way, the FIMs for individual poses in \eqref{eq:cost_info} need to be computed multiple times for different motion samples or the iterations in optimization, which is the computational bottleneck for solving \eqref{eq:active_loc_full}.
Specifically, the calculation of $\I_{\T_i}$ requires iterating all the landmarks in the environment and evaluating the individual FIM for all the visible ones (the sum in \eqref{eq:cost_info}), which scales linearly with the number of landmarks.
Moreover, $\I_{\T_{i}}$ needs to be recomputed from scratch once the pose $\T_i$ changes (both the visibility and the Jacobian in \eqref{eq:fim_gaussian} change), which introduces redundant computation across the iterations in the planning algorithm as well as multiple planning sessions.
This motivates us to look for an alternative formulation of \eqref{eq:cost_info} to mitigate the bottleneck.

It is worth mentioning that, compared with complete probabilistic treatment as in \cite{Burgard97ijcai, Roy99icra}, we make the simplification in the problem formulation \eqref{eq:active_loc_full} \eqref{eq:cost_info} that the localization process purely depends on the measurements (\ie no prior from the past).
However, this is not a limitation of our work.
The computational bottleneck exists as long as the Fisher information is used to characterize the visual estimation process.
The essence of this work is a compact representation of the information to allow efficient computation, which is widely applicable.

%% file: sections/approx_fim.tex
\section{Approximating FIM: Factoring out the Rotation}
\label{sec:fim_approx}

In this section, we focus on the formulation of the Fisher information for a single pose, since the FIMs of different poses are calculated independently in the same way.
Let $\Twc = \{\Rwc, \twc\}$ stand for the pose of the camera in the world frame.
The first step of computing the FIM at $\Twc$ is to identify the landmarks that can be matched in the image taken at the pose.
As shown in \cite{Alcantarilla10icra,Alcantarilla11icra}, this depends on various factors and is a non-trivial task.
In this work, we decompose this process into two steps.
First, we select the set of landmarks $V_{\twc} = \{\pw_i\}_{i=1}^{N} $ that can be observed from the position $\twc$, which needs to be determined by considering occlusion and viewpoint change (see an example in \Sec\ref{subsec:planning}).
Second, once $V_{\twc}$ is known, the remaining factor to consider is whether the landmarks are within the FoV of the camera, which we refer to as ``visibility'' in the rest of the paper.
We model this factor using an indicator function as shown below.
The motivation of this separation is that selecting $V_{\twc}$ considers the factors that \emph{only} depend on $\twc$ and thus can be precomputed for each voxel in a voxel grid, as desired in our map representation.

Given $V_{\twc} =  \{\pw_i\}_{i=1}^{N} $, we can write the FIM for the pose of interest as
\begin{eqnarray}
\I_{\Twc} = \sum_{i=1}^{N}v(\Twc, \pw_i)\I_i,
\label{eq:vis_fim}
\end{eqnarray}
where $v(\Twc, \pw_i)$ is a binary valued function indicating the visibility of the $i$th landmark due to the limited FoV, and $\I_i$ stands for the FIM corresponding to the observation of the $i$th landmark.
Conceptually, our goal is to find an approximation $\S(\Twc, \pw_i) \approx v(\Twc,\pw_i)\I_i$ that can be written as $\S(\Twc, \pw_i) = \S(\H(\twc, \pw_i), \Rwc)$ and satisfies
\begin{equation}
\I_{\Twc}\approx \sum_{i=1}^{N}\S(\Twc, \pw_i)
= \S(\sum_{i=1}^{N}\H(\twc, \pw_i), \Rwc).
\label{eq:info_separation}
\end{equation}
In words, we would like to find an approximation that can be factored into two components, one of which does not depend on rotation (\ie $\H(\cdot)$ in \eqref{eq:info_separation}), and the approximation is linear in terms of the rotation-independent part.

The linear form lead to two favorable properties.
First, for one position $\twc$, the sum of the rotation-independent $\H(\cdot)$ of all the landmarks need to be computed \textit{only once}, and the sum can be used to calculate the approximated information at this position for arbitrary rotations in constant time;
second, we can easily update the sum when new landmarks are added or old ones deleted.
This form naturally leads to a volumetric representation (\ie pre-compute and store values of interest in a voxel grid) that allows online update, as described in \Sec\ref{subsec:fif_map}.
The approximation \eqref{eq:info_separation} is achieved by first carefully parameterizing the information matrix $\I_i$ to be rotation-invariant (\Sec\ref{subsec:rot_inv_fim}) and replacing the binary valued function $v(\cdot)$ with smooth alternatives (\Sec\ref{subsec:vis_approx}).

\subsection{Rotation Invariant FIM}
\label{subsec:rot_inv_fim}
The observation of a 3D landmark $\pw$ can be represented in different forms, such as (normalized) pixel coordinates and bearing vectors.
In this work, we choose to use the bearing vector $\f$ because of its ability to model arbitrary FoVs \cite{Zhang16icra}.
Then the noise-free measurement model of a landmark $\pw_i$ is
\begin{eqnarray}
\f_i = \frac{\pc_i}{\norm{\pc_i}_2}=\frac{1}{n_i}\pc_i,\quad \pc_i = \Tcw\pw_i,
\end{eqnarray}
and the Jacobian of interest is
\begin{eqnarray}
\J_i = \frac{\partial\f_i}{\partial\pc_i} \frac{\partial\pc_i}{\partial\Twc}.
\label{eq:jac_f}
\end{eqnarray}
While the first part in \eqref{eq:jac_f} is trivially 
\begin{equation}
\frac{\partial\f_i}{\partial\pc_i} = \frac{1}{n_i}\Id_{3} - \frac{1}{n^3_i}\pc_i(\pc_i)^{\top},
\end{equation}
the derivative $\frac{\partial\pc_i}{\partial\Twc}$ is more involved.
To handle the derivatives related to 6 DoF poses without over-parametrization, the element in $\se(3)$ (denoted as $\secoords$) is often used. In our case, $\frac{\partial\pc_i}{\partial\Twc}$ is replaced by
\begin{eqnarray}
\frac{\partial\pc_i}{\partial\Twc} \rightarrow 
\frac{\partial(\expm(\secoords^{\wedge})\Twc)^{-1}\pw_i}{\partial\secoords}
\;\; \text{or} \;\;
\frac{\partial(\Twc\expm(\secoords^{\wedge}))^{-1}\pw_i}{\partial\secoords},
\label{eq:deriv_se3}
\end{eqnarray}
where $\expm(\cdot) $ is the exponential map of the Special Euclidean group $\SE(3)$.
The two forms correspond to expressing the perturbation $\delta\secoords$ globally in the world frame or locally in the camera frame respectively.
Using the first form, we have the Jacobian in \eqref{eq:jac_f} as
\vspace{-0.5ex}
\begin{eqnarray}
\J_i = \frac{\partial\f_i}{\partial\pc_i}\Rcw[-\Id_{3},\quad \skewsym{\pw_i}].
\label{eq:jac_pose_global}
\end{eqnarray}
With the global perturbation formulation, for two poses that differ by a relative rotation $\Twc$ and $\T_{\texttt{wc'}} = \{\Rwc\R_{\texttt{cc'}}, \twc\}$, their Jacobians \eqref{eq:jac_pose_global} have a simple relation
$
\J_i^{\prime} = \R_{\texttt{c'c}} \J_i
$,
from which the corresponding FIMs turn out to be the same
\vspace{-0.5ex}
\begin{eqnarray}
\I^{\prime}_i \overset{\eqref{eq:fim_gaussian}}{=}
\frac{1}{\sigma^2}\J_i^{\top}\R_{\texttt{cc'}}\R_{\texttt{c'c}} \J_i= \I_i.
\label{eq:global_fim_eq}
\end{eqnarray}

The rotation-invariance is not surprising.
Intuitively, since we are considering only part of \eqref{eq:vis_fim} (without visibility constraint) and modeling the camera as a general bearing sensor, the camera should receive the same information regardless of its rotation.
Moreover, the choice of global frame expresses the constant information in a fixed frame, resulting in the invariance \eqref{eq:global_fim_eq}.
If the local perturbation in \eqref{eq:deriv_se3} is chosen, such invariance is not possible, and the information matrix will be related by an adjoint map of $\SE(3)$ \cite[Ch.~2]{Strasdat12thesis}.

To summarize, by choosing the bearing vector as the observation and parameterizing the pose perturbation in the global frame, the information matrix, without the visibility constratint, is rotation-invariant.
Next, we will see how to handle the visibility function $v(\cdot)$ in \eqref{eq:vis_fim}.

\begin{figure}[t]
  \centering
  \begin{subfigure}[]{0.42\linewidth}
    \includegraphics[width=\linewidth]{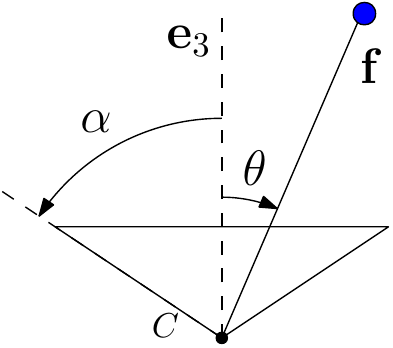}
    \caption{Visibility as a function of $\theta$}
    \label{fig:theta_vis}
  \end{subfigure}
  \begin{subfigure}[]{0.52\linewidth}
    \includegraphics[width=\linewidth]{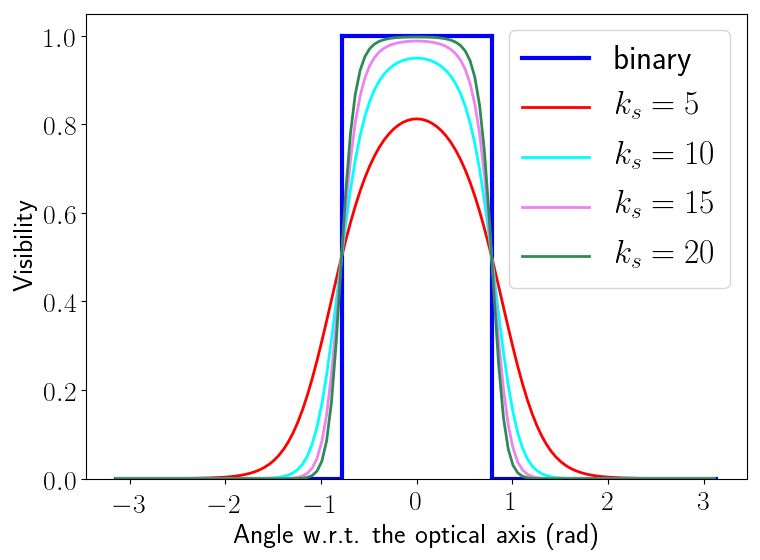}
    \caption{Visibility approximated by sigmoid functions with different $k_s$}
    \label{fig:sigmoid_vis}
  \end{subfigure}
  \caption{
    Visibility modeling.
   $\alpha$ is half of the FoV, $\f$ is the bearing vector observation, $\e_3$ is the optical axis of the camera, and $C$ is the projection center.
  }
  \label{fig:vis_approx}
\end{figure}

\subsection{Visibility Approximation}
\label{subsec:vis_approx}

The exact visibility $v(\Twc, \pw)$ is a non-trivial function, as the horizontal/vertical/diagonal FoVs are not the same.
In practice, to check whether a point is visible at a pose, one needs to project the 3D point to the image plane $\px = \proj(\Twc, \pw)$ and check whether the projected pixel coordinates $\px$ is within the image boundary $I$:
\begin{equation}
  v(\Twc, \pt^{\text{w}}) =
  \begin{cases}
    1,  & \px \in I \\
    0,  & \px \notin I 
  \end{cases},
  \label{eq:exact_vis_func}
\end{equation}
For simplicity, we assume that the visibility $v(\cdot)$ is a function of the angle $\theta$ between the bearing vector $\f$ of the landmark and the optical axis $\e_3 = [0, 0, 1]^{\top}$ in the camera frame:
\begin{equation}
  v(\Twc, \pt^{\text{w}}) \approx v(\theta) =
  \begin{cases}
    1,  & \theta \leq \alpha \\
    0,  & \theta > \alpha
  \end{cases},
  \label{eq:theta_vis_func}
\end{equation}
where $\alpha$ is half of the camera FoV.
This essentially assumes that the FoVs along different directions of the image plane are the same, as illustrated in \Fig\ref{fig:theta_vis}.
Since our goal is to arrive at the form \eqref{eq:info_separation}, we further assume that the simplified $v(\theta)$ can be written, by certain approximation, as a dot product of two vectors:
\begin{equation}
  v(\theta) \approx (\vr(\Rwc))^{\top} \vp(\twc, \pw),
  \label{eq:general_vis_approx}
\end{equation}
where $\vr$ and $\vp$ only depend on the rotation and position respectively.
We keep only the parameters that are related to $\Twc$ and $\pw$ for brevity.
The motivation of this form is for the easy separation of the rotation-dependent and translation-dependent components.
Once we have an approximation that satisfies \eqref{eq:general_vis_approx} with $\vr$ and $\vp$ of length $N_v$, the full FIM from $N$ landmarks \eqref{eq:vis_fim} can be written in the form of \eqref{eq:info_separation}:
\begin{equation}
\begin{aligned}
\I_{\Twc} &\overset{\eqref{eq:general_vis_approx}}{\approx}
\sum_{i=1}^{N} (\vr)^{\top} \vp_i \I_i
= \sum_{i=1}^{N} \diag_6((\vr)^{\top} \vp_i) \I_i \\
&=  \diag_6((\vr)^{\top}) \sum_{i=1}^{N} \diag_6(\vp_i) \I_i \\
&= \V_{\I}(\Rwc) \C_{\I}(\twc, \{\pw_i\}_{i=1}^{N}),
\label{eq:fim_separation}
\end{aligned}
\end{equation}
where
\begin{eqnarray}
  \V_{\I}(\Rwc) &\triangleq& \diag_6((\vr)^{\top}),
  \label{eq:general_info_query}
  \\
  \C_{\I}(\twc, \{\pw_i\}_{i=1}^{N}) &\triangleq& \sum_{i=1}^{N}\diag_6(\vp_i) \I_i.
  \label{eq:general_info_pif}
\end{eqnarray}
$\diag_n(\A)$ denotes a diagonal matrix by repeating $\A$ by $n$ times on the diagonal, 
$\V_{\I}(\cdot)$ is a $6\times 6 N_v$ matrix and only depends on the rotational component of the pose,
and $\C_{\I}(\cdot)$ is of size $6N_v \times 6$ and only depends on the positions of the camera and the landmarks.
We refer to $\C_{\I}(\cdot)$ as the \emph{positional information factor} at $\twc$ in the rest of the paper.

Our goal now is to find visibility approximations that satisfies the form of a vector dot product $\eqref{eq:general_vis_approx}$, which allows the desired separation in \eqref{eq:fim_separation}.
There are possibly different ways of achieving this.
Next, we first present two approaches that we explore and validate in this paper and then discuss the possibility of using alternative models.

\begin{figure}[t]
    \includegraphics[width=\linewidth]{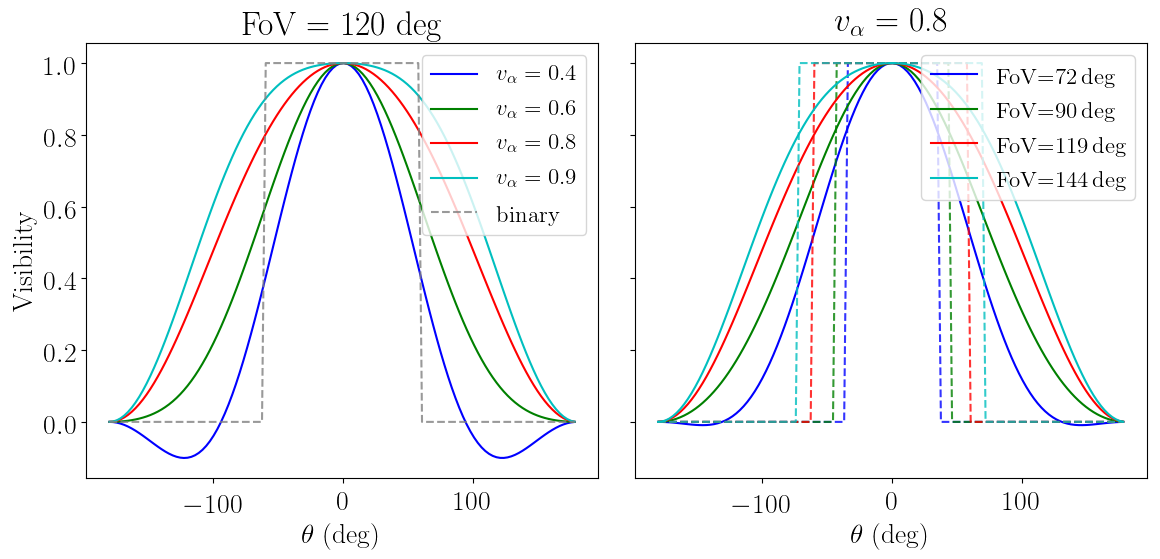}
    \caption{
      Examples of the quadratic visibility approximations \eqref{eq:quad_vis_approx} that satisfy different conditions in \eqref{eq:quadratic_conditions}.
      \textbf{Left:} Different visibility at the FoV boundary for a fixed FoV.
      \textbf{Right:} Different FoV for a fixed value at the FoV boundary.
      The dashed lines correspond to the binary visibility function \eqref{eq:theta_vis_func}.
      }
  \label{fig:quad_approx_eg}
\end{figure}

\subsubsection{Quadratic}
In the previous conference version \cite{Zhang19icra}, we used a quadratic function of $\cos\theta$ to approximate the simplified visibility $v_{\qd}(\theta) \approx v(\theta)$:
\begin{equation}
  v_{\qd}(\theta) = k_2 \cos^{2}\theta + k_{1} \cos^{}\theta + k_0,
  \label{eq:quad_vis_approx}
\end{equation}
where $\cos\theta= (\Rwc\e_3)^{\top} (\frac{\pt^0}{\lVert\pt^0\rVert_2})$, and $\pt^0 = \pw - \twc$.
With $\Rwc\e_3 = [z_1, z_2, z_3]$ and ${\pt^0}/{\lVert\pt^0\rVert_2} = [p_1, p_2, p_3]$,
\eqref{eq:quad_vis_approx} can be written as:
\begin{equation}
  \begin{aligned}
  v_{\qd}(\theta) &= \sum_{p=0}^{p=2} k_p (z_1p_1 + z_2p_2 + z_3p_3)^{p} \\
  &= (\vr(z_1, z_2, z_3))^{\top} \vp(p_1, p_2, p_3),
  \end{aligned}
  \label{eq:quad_vis_component}
\end{equation}
where
\begin{equation}
  \begin{aligned}
    \vr = &[k_2z_1^2, k_2z_2^2, k_2z_3^2, 2k_2z_1z_2, 2k_2z_1z_3, 2k_2z_2z_3, \\
    &k_1z_1, k_1z_2, k_1z_3, k_0]\\
    \vp = &[p_1^2, p_2^2, p_3^2, p_1p_2, p_1p_3, p_2p_3, p_1, p_2, p_3, 1]\\
  \end{aligned}
\end{equation}
\eqref{eq:quad_vis_component} satisfies the form \eqref{eq:general_vis_approx}, and the length of $\vr$ and $\vp$ is $10$.
To determine the coefficients, we specify the visibility at $3$ angles and solve for $\{k_2, k_1, k_0\}$ accordingly.
In particular, the values at $\theta = 0$ (center of the camera FoV), 
$\theta = \pi$ (behind the camera)
and at the boundary of the camera FoV ($\theta=\alpha$ as in \Fig\ref{fig:vis_approx}) are set:
\begin{equation}
  v_{\qd}(0) = 1.0,\quad v_{\qd}(\pi) = 0.0, \quad v_{\qd}(\alpha) = v_{\alpha}
  \label{eq:quadratic_conditions}
\end{equation}
Several examples for different $\alpha$ and $v_\alpha$ are shown in \Fig\ref{fig:quad_approx_eg}.

The quadratic approximation is simple and can adapt to different FoVs to a certain extent.
However, its expressive power is rather limited.
As seen in \Fig\ref{fig:quad_approx_eg}, to accommodate for different FoVs, $v_{\qd}(\theta)$ can be non-monotonic with respect to the view angle and have negative values.
In addition, there is often a heavy tail for large $|\theta|$.

\subsubsection{GP Regression}
\label{subsec:gp_reg}
Given a landmark $\pw$, we can also use a GP to regress the visibility of it in a camera pose $\Twc$.
Since we simplify the visibility as \eqref{eq:theta_vis_func}, the visibility of a landmark does not depend on the full camera rotation $\Rwc$ but the direction of the optical axis $\z = \Rwc\e_3$,
which we use as the input for the GP.
In particular, we design the GP to approximate a smoother version of \eqref{eq:theta_vis_func}
\begin{equation}
  v_{\text{sig}}(\theta) = \frac{1}{1+e^{-k_s (\cos\theta - \cos\alpha)}}
  \label{eq:sigmoid_vis_func}
\end{equation}
The function is illustrated in \Fig\ref{fig:sigmoid_vis} for different $k_s$, and $k_s = 15$ is used in the rest of the paper.
First, we sample $N_s$ poses with the same position as $\Twc$ but different rotations $R^s = \{\R_{\text{wc}, g}\}_{g=1}^{N_s}$.
The corresponding orientations of the optical axis are denoted as $Z^s = \{\z^{s}_{g}\}_{g=1}^{N_s}$.
We then compute the visibility of the landmark at these poses according to \eqref{eq:sigmoid_vis_func}, denoted as $\mathbf{v}^s = [v_1, v_2, \ldots, v_{N_s}]^{\top}$.
Given a kernel function defined for two unit 3D vectors $k(\z_i, \z_j)$, the visibility of the landmark at $\Twc$ can be approximated (interpolated) as the regressed mean value from GP $ v(\theta) \approx v_{\gp}(\z)$:
\begin{equation}
v_{\gp}(\z) = \K_{\z} \K_{\z}^{s} \mathbf{v}^{s},
\label{eq:gp_vis_approx}
\end{equation}
where $\K_{\z}$ and $\K_{\z}^s$ are of $1 \times N_s$ and $N_s \times N_s$, and 
\begin{equation}
  \K_{\z}(1, g) = k(\z, \z_g^{s}),\;
  \K_{\z}^{s}(g_i, g_j) = k(\z_{g_i}^{s}, \z_{g_j}^{s}),
\end{equation}
for $g, g_i, g_j \in [1, 2, \ldots, N_s]$.
It  can be seen that \eqref{eq:gp_vis_approx} satisfies the form \eqref{eq:general_vis_approx} by observing $\vr = \K_\z$
and $\vp = \K_{\z}^s \mathbf{v}^s$,
where $\K_\z$ only depends on the $\Rwc$ and the sampled rotations,
and $\K_{\z}^s$ only depends on the sampled rotations, $\twc$ and $\pw$.
The length of $\vr$ and $\vp$ is the same as the number of the samples $N_v = N_s$.

Since our goal is to have a collective term for different landmarks, the same sampled rotations are used for each landmark to have the same $\K_\z$ to allow \eqref{eq:fim_separation}.
As for the kernel function, we use the squared exponential function, adapted for 3D vectors 
$
  k_{\text{SE}}(\z_1, \z_2) =  \sigma^2 \exp(-\frac{\lVert\z_1 - \z_2\rVert_2^2}{2l^2}),
$
Regarding the hyperparameters, the noise parameter is fixed to $\sigma^2 = 1e-10$, and the length scale $l$ is optimized following the standard approach of maximizing the marginal likelihood \cite[p.~112]{Rasmussen05gpml}.
The training data is generated by calculating the simplified visibility \eqref{eq:sigmoid_vis_func} at the sampled rotations for a set of random landmarks.

\begin{figure}[t]
    \includegraphics[width=0.9\linewidth]{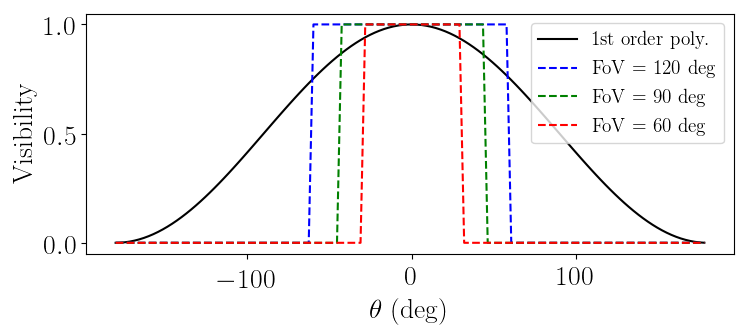}
    \caption{
        Using the first order polynomial ($p=1$ in \eqref{eq:poly_vis_approx}) for visibility approximation.
        With the visibility being 1 and 0 at the center of the FoV and behind the camera respectively, the first order polynomial approximation (solid line) is fully determined and cannot adapt to the visibility $v(\theta)$ of different FoVs (dashed lines), as opposed to the quadratic model in \Fig\ref{fig:quad_approx_eg}.
      }
  \label{fig:1st_order_poly}
\end{figure}
\subsubsection{The Choice of Visibility Models}
\label{ssub:vis_model_choice}
To arrive at the general form \eqref{eq:info_separation}, we choose to approximate the visibility with the dot product of two vectors \eqref{eq:general_vis_approx}, which is probably not the only viable option.
Furthermore, to satisfy the dot product form, there are also other models that can be used in addition to the quadratic and GP approximations described above.
While it is not possible to exhaustively list the alternatives, we discuss next the feasibility of some variations of the two approximations to put our choices in context.
In our experiment, we will instead focus on evaluating the quadratic and GP approximations.

The quadratic approximation is a special case of a polynomial approximation of an arbitrary order $p$.
In particular, \eqref{eq:quad_vis_approx} can be extended as
\begin{equation}
  v_{\poly}(\theta) = k_p \cos^{p}\theta + \cdots + k_{1} \cos^{}\theta + k_0,
  \label{eq:poly_vis_approx}
\end{equation}
and we can separate the bearing vector $\z$ and $\pt^0$ in the same way as \eqref{eq:quad_vis_component} (though more complicated for higher order).
The coefficients can be found via fitting \eqref{eq:poly_vis_approx} into \eqref{eq:theta_vis_func} by, for example, least squares regression and enforcing \eqref{eq:poly_vis_approx} to be a monotonic function with respect to $|\theta|$.
It can be expected that increasing the order $p$ will reduce the approximation error but increase the length of $\vr$ and $\vp$ and thus the computational cost.
We choose $p=2$ due to its efficiency and relatively good expressive power.
For $p=1$, \eqref{eq:poly_vis_approx} becomes a linear function $k_1\cos\theta + k_0$ with only two parameters and has very limited fitting capability.
For example, by enforcing the first two constraints in \eqref{eq:quadratic_conditions} in the linear function, the coefficients are fully determined as $k_0 = k_1 = 0.5$, leaving no possibility of adapting to different FoVs, as shown in \Fig\ref{fig:1st_order_poly}.

The GP approximation $\eqref{eq:gp_vis_approx}$ essentially regresses the visibility of a landmark from the camera optical axis direction $\z$.
In terms of satisfying $\eqref{eq:general_vis_approx}$, GP is not strictly needed.
As a simple example, we can use a linear function of $\z = [z_1, z_2, z_3]$:
\begin{equation}
  v_{\lin}(\z) = \z^{\top} \mathbf{l} + b 
  = \underbrace{[z_1, z_2, z_3, 1]^{\top}[l_1, l_2, l_3, b]}_{\text{\small satisfies \eqref{eq:general_vis_approx}}},
  \label{eq:lin_z_regress}
\end{equation}
where $\mathbf{l} = [l_1, l_2, l_3]$ is a $3\times1$ vector, and $b$ a scalar.
The approximation is similarly limited in terms of the expressive power as the $1$st order polynomial mentioned above.
\eqref{eq:lin_z_regress} can be further extended to a linear combination of arbitrary integer exponents of $[z_1, z_2, z_3]$, but it is not clear which exponents should be chosen for better approximation.
We thus choose GP with the squared exponential kernel due to its expressive power and flexibility.
The squared exponential kernel is universal \cite{micchelli06jmlr} and is the de-facto standard kernel for GP.
Moreover, adding more samples naturally increases the approximation accuracy without introducing additional complexity in implementation.

\subsection{Discussion}
\label{subsec:fim_approx_discussion}

\PAR{Discrepancy in Visibility Approximation}
We arrive at the convenient form \eqref{eq:info_separation} at the cost of introducing discrepancy between the exact visibility model and the visibility approximations.
The discrepancy consists of two parts: 1) the difference between the simplified visibility function $\eqref{eq:theta_vis_func}$ and the exact one $\eqref{eq:exact_vis_func}$ 2) the difference between the (linearly) separable visibility models \eqref{eq:quad_vis_approx} \eqref{eq:gp_vis_approx} and the simplified model \eqref{eq:theta_vis_func}.
We will study in details about the impact of the discrepancy via simulation in \Sec\ref{subsec:sim}.

\PAR{GP and Spherical Interpolation}
The GP visibility approximation in \eqref{eq:gp_vis_approx} is a model that regressed the scalar visibility over an sphere (\ie unit vectors for optical axis orientations).
In terms of GP regression, it is essentially a weighted sum (weights determined by the kernel function) of the values at different samples (\eg $\mathbf{v}^s$ in \eqref{eq:gp_vis_approx}) and thus is often seen as an interpolation method (also known as \emph{kriging}).
Therefore, the aforementioned GP visibility approximation is conceptually similar to spherical interpolation (see \eg \cite{Robeson97cgis, Carfora07jcam}).

To summarize, in this section, we described our formulation of the FIM that satisfies the form \eqref{eq:info_separation}, achieved by identifying the rotational-invariant component and approximating the visibility function.
Next, we will describe how to design a map representation for perception-aware planning using the proposed formulation.

%% file: sections/fif_planning.tex
\section{Building a Map for Perception-aware Planning}
\label{sec:fif_planning}
\subsection{The Fisher Information Field}
\label{subsec:fif_map}
\subsubsection{Representation, Query and Update}
Using the formulation  \eqref{eq:fim_separation}, \eqref{eq:general_info_query} and \eqref{eq:general_info_pif}, we propose a volumetric representation, namely the \textit{Fisher Information Field} (FIF), for perception-aware planning.
In particular, after discretizing the space of interest into voxels, we compute the positional factors $\C_{\I}(\cdot)$ at the centers of the voxels (from all the 3D landmark), and store each factor in the corresponding voxel.
Then, when the information of a certain pose is queried, 
the related positional factors (by nearest neighbor or interpolation) are retrieved, 
and \eqref{eq:fim_separation} is used to recover the information in constant time.
The method is illustrated in \Fig\ref{fig:vox_mapping}.
Once the field is built,
the query of the information for an arbitrary pose only requires a linear operation of the related positional factors
instead of checking all the points in the point cloud,
which is the key advantage of the proposed method.

\PAR{Offline Processing vs. Online Update}
The most straightforward use case of FIF is to build a map offline for a known environment, and use it for online (re-)planning.
Note that determining whether a landmark can be used for localization, as we mentioned at the beginning of \Sec\ref{sec:fim_approx}, is usually a complicated process in non-trivial environments (see \cite{Alcantarilla10icra, Alcantarilla11icra} and \Sec\ref{subsec:planning}).
It is coupled with the detailed scene structure (\eg occlusion), viewpoint change with respect to the images used to build the map of sparse landmarks, and the method for establishing the correspondences (\eg feature-based or direct methods).
Therefore, calculating FIF from a known map is suitable for offline processing, where sophisticated and expensive methods can be used to determine whether the landmarks can be observed from the positions of interest.
On the other hand, it would be very useful to be able to build such a map incrementally online, for example, during the exploration of a previous unknown environment.
This is in principle trivial in terms of our map representation.
Since the positional information factors \eqref{eq:general_info_pif} is in the form of the summation of components calculated from each landmark independently, 
adding/deleting the contribution of a landmark can be done by adding/subtracting the corresponding components from existing factors.
Updating the contribution of a landmark (\eg map update in SLAM) is also possible by computing the difference of the component corresponding to the landmark and updating the related positional factors.
The main hurdle, however, is how to reliably and efficiently determine the matchability of landmarks online, especially with only partial information of the environment.
While it is possible for scenes with simple layouts (see \Sec\ref{subsec:inc_update}), it is not clear how it can be done in a principled manner in general.
Since the main focus of our work is a novel representation for efficient planning, we focus on the use case in an known environment in the rest of the paper.

\subsubsection{Memory Complexity and Trace Factor}
The constant query time comes at the cost of extra memory.
In particular, the positional information factor $\C_{\I}(\cdot)$ at each location consists of $36N_v$ floats, where $N_v$ is the length of vector in the dot product approximation \eqref{eq:general_vis_approx}.
Admittedly, the size of storage needed is non-negligible (\eg 360 float numbers for quadratic visibility model and 1800 for GP with 50 samples), and it increases linearly with the number of voxels in the field.
But the memory footprint is still acceptable in practice, as we will show in \Sec\ref{subsec:planning} in a realistic setup.

Note that the aforementioned information representation can be used to recover the full approximated information matrices ($6\times6$).
However, in the overall cost for planning \eqref{eq:cost_info}, only one scalar metric $s(\cdot)$ is needed.
This brings the possibility of reducing the memory usage by directly expressing one specific metric instead of the full information matrix.
Out of different metrics often used with the Fisher information \cite[Ch.~6\&9]{Friedrich06book}, the T-optimality criterion, which is the matrix trace, is especially suitable (\ie a linear function) for this purpose.
In particular, taking the trace of the approximated information \eqref{eq:fim_separation}, we can arrive at the following form
\begin{equation}
\Tr(\I_{\Twc}) \approx \sum_{i=1}^{N} (\vr)^{\top}\vp_i \Tr(\I_i) =
\V_{\Tr}(\Rwc) \C_{\Tr}(\twc, \{\pw_{i}\}_{i=1}^{N})
\label{eq:fim_trace_separation}
\end{equation}
where
\begin{eqnarray}
  \V_{\Tr}(\Rwc) &\triangleq& (\vr)^{\top},\\
  \label{eq:general_trace_query}
  \C_{\Tr}(\twc, \{\pw_i\}_{i=1}^{N}) &\triangleq& \sum_{i=1}^{N}\vp_i\Tr(\I_i).
  \label{eq:general_trace_pif}
\end{eqnarray}
Similarly, we call \eqref{eq:general_trace_pif} the \emph{positional trace factor}.
Notably, $\V_{\Tr}(\cdot)$ and $\C_{\Tr}(\cdot)$ are of size $1 \times N_v$ and $N_v \times 1$ respectively, reducing the memory usage by a factor of $36$ compared with the information factor \eqref{eq:general_info_pif}.
The trace factor can be used in the same volumetric mapping framework mentioned above, but only requires $N_v$ float numbers for one voxel (at the cost of losing certain information contained in the full FIMs).

\subsection{Integration in Motion Planning}
\label{subsec:fif_motion_planning}
Conceptually, integrating the FIM in motion planning is straightforward.
For instance, we can define a certain value as the threshold to determine whether a pose can be localized, or we can add different metrics of the FIM as an additional term in the cost function, as described in \eqref{eq:active_loc_full}.
However, there are several problems with the naive approach.
First, the thresholds are less intuitive to choose (\eg than the distance threshold for collision avoidance).
Second, since different information metrics for the same FIM have very different values, the thresholds or weights for these information metrics have to be chosen separately, which makes the parameter tuning complicated.
More importantly, this also makes a fair comparison difficult.
For example, if the thresholds/weights using two metrics are chosen differently, how can we tell a worse performance is due to the metric or the lack of tuning?
The same problem exists for using different types of FIFs as well, since the information metrics from them are also different due to approximation.
To overcome the above problems, we propose a unified approach of defining thresholds or costs for the metrics from the FIM.
Instead of defining the thresholds for the information metrics directly, we compute the thresholds from certain specifications of the landmarks.

\PAR{Information Threshold}
For a certain pose, we assume that if there are $M$ landmarks in the camera FoV that are within $d_{\min}$ to $d_{\max}$ distance to the camera, the pose is considered to be able to be localized.
We first randomly generate several sets of landmarks that satisfy the criteria.
Then, for a certain information metric from a certain map representation (FIF or the point cloud), we first construct the map representation if necessary, calculate the information metric from the map representation and use the average value as the threshold for this combination of information metric and map representation.
Therefore, given the \emph{same} landmark specifications ($M$, $d_{\min}$ and $d_{\max}$), the thresholds for different combinations of information metrics and map representations are calculated automatically and correspond to the same physical meaning.
The randomly generated landmarks could, in theory, form degenerate configurations for visual localization, which we didn't encounter in our experiments.
Such situation could be easily avoided by checking the rank/condition number of the FIM and excluding the corresponding sample from the average, if it becomes a problem in practice.

\begin{figure}[t]
  \centering
  \includegraphics[width=0.7\linewidth]{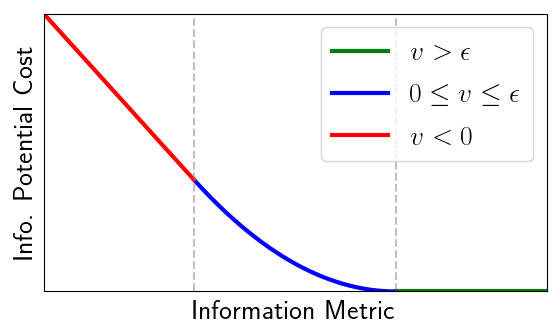}
  \caption{Illustration of the information potential cost in \eqref{eq:info_pot_cost}.}
  \label{fig:info_potential}
\end{figure}

\PAR{Information Potential Cost}
For optimization-based methods, we further define an information potential cost similar to the collision potential cost in \cite{Zucker13ijrr}.
The intuition is that we no longer care about the improvement of the localization quality after it has reached a certain level.
Specifically, assuming that the pose can be estimated well enough with $M$ landmarks in the FoV from $d_{\min}$ to $d_{\max}$ distance to the camera, we first calculate the threshold $\epsilon_{\fim}$ as mentioned before.
We then define the information potential cost $c_{\fim}$ as
\begin{equation}
  c_{\fim}(v_{\fim}) = 
  \begin{cases}
    0, & v_{\fim} > \epsilon_{\fim}\\
    k_q (v_{\fim} - c_{\fim})^2, &  0 \leq v_{\fim} \leq \epsilon_{\fim} \\
    k_l \cdot v_{\fim} + b_l,  & v_{\fim} <= 0
  \end{cases}
  \label{eq:info_pot_cost}
\end{equation}
where $v_{\fim}$ is the information metric.
$k_q > 0$ is chosen empirically.
$k_l$ and $b_l$ are calculated to guarantee the continuity in both $c_{\fim}$ and its derivative at $c_{\fim} = 0$ after $k_q$ has been chosen.
An illustration of the information potential cost is shown in \Fig\ref{fig:info_potential}.

With our approach, we can specify the thresholds for different metrics/map representations in the same way and, in optimization-based motion planning, use the same weight for the cost related to the FIM.
This greatly simplifies the experiments in \Sec\ref{subsec:planning} and makes the results using different information metrics and map representations comparable.

%% file: sections/experiments.tex
\section{Experiments}
\label{sec:exp}
We implemented the proposed Fisher Information Field in C++. We used the library from \cite{Oleynikova17iros} for the voxel hashing functionality. 
Next, we present both quantitative and qualitative results using our implementation.
Specifically, we aim to answer the following questions:
\begin{itemize}
  \item How do different visibility approximations affect the efficiency and accuracy of FIF?
  \item How can FIF be used with different motion planning algorithms? What are the benefits?
  \item How does FIF compare with point clouds in terms of the computation of the FIM and perception-aware planning?
\end{itemize}
Moreover, we also present qualitative results of building FIF from the output of a visual-inertial odometry pipeline incrementally, showing the possibility of using our method in previous unknown environments.

\begin{table*}[t]
  \centering
  \begin{tabular}{ l l c c c c c c c c c}
    \toprule
      & & PC & Q-I & Q-T & G-I-30 & G-I-70 & G-I-150 & G-T-30 & G-T-70 & G-T-150 \\
    \midrule
    \multirow{2}{*}{Build} &
    $t_\text{build}$ (sec) & - & 7.34  & 2.96 & 17.84  & 41.10  & 124.74  & 10.93  & 26.48  & 82.72   \\
    &
    Mem. (MB) & 0.02  & 58.50 & 1.62 & 135.00 & 315.00 & 675.00  & 3.75   & 8.75   & 18.75   \\
    \midrule
    \midrule
    \multirowcell{4}{Query  \\(\si{us})}&
    FIM & 97.3 & 0.4 & - & 0.9 & 2.7 & 4.7 & - & - & -  \\
    & 
    $\det$  & 98.2 & 3.0 & - & 5.7 & 12.1 & 24.9 & - & - & - \\
    &
    $\mineig$ & 102.9 & 6.3 & - & 8.6 & 14.7 & 27.6 & - & - & - \\ 
    &
    $\text{Trace}$ & 97.7 & 1.6 & 0.6 & 3.9 & 10.0 & 23.3 & 1.1 & 1.9 & 3.4 \\
    \bottomrule
  \end{tabular}
  \caption{Time and memory required for building different types of FIF, and the time per query for full FIM, determinant ($\det$), smallest eigenvalue ($\mineig$) and the trace of FIM (\trace).
    By pre-computation, different types of FIF are significantly faster than using point clouds (PC) at query time, which is important for online applications such as motion planning.
    Note that querying $\det$, $\lambda_{\text{min}}$ and $\trace$ from different FIFs uses bi-linear interpolation.
  }.
  \label{tb:sim_complexity}
\end{table*}

\begin{figure}[t]
  \centering
  \includegraphics[width=0.8\linewidth]{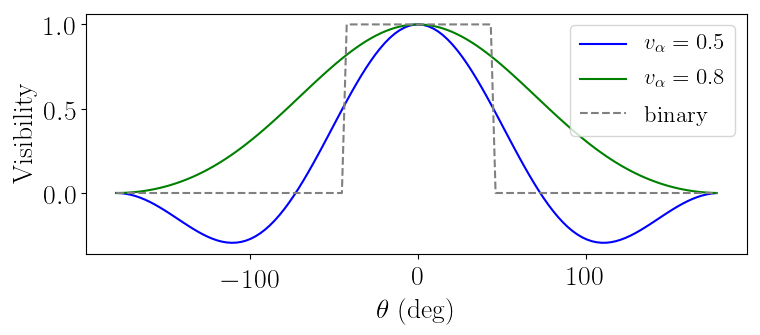}
  \caption{
    The quadratic visibility approximations tested in the simulation.
    The FoV is $90 \deg$.
    The negative values for $v_\alpha=0.5$ may seem unreasonable, but this behavior essentially punishes the situation where landmarks are outside the FoV, which is desirable intuitively.
  }
  \label{fig:sim_quad_vis}
\end{figure}

\begin{table}[t]
  \centering
  \scriptsize{
  \begin{tabular}{l p{0.6cm} p{0.6cm} p{0.6cm} p{0.6cm} p{0.6cm} p{0.6cm} p{0.6cm}}
    \toprule
     & {Q-I-0.5} &{Q-I-0.8}   
     & {G-I-30}  & {G-I-50} 
     & {G-I-70} 
     & {G-I-120} & {G-I-150}  \\
    \midrule
    {Diff. (\%)} 
    & 64.26 & 164.92 
    &  11.15 & 9.88 
    & 9.49 
    & 9.28 & 9.45 \\
    \bottomrule
  \end{tabular}
  }
  \caption{Relative difference with respect to the exact FIM (computed from the point cloud and the exact pinhole camera model) in terms of the Frobenius norm (see \eqref{eq:fim_rel_diff}).}
  \label{tb:sim_fim_acc}
\end{table}

\subsection{Simulation}
\label{subsec:sim}
As mentioned in \Sec\ref{subsec:fif_map}, constructing FIF requires the ability to determine which landmarks can be matched from a certain viewpoint and is a difficult task in non-trivial setups. 
Since the focus of the paper is a representation for the FIM that allow efficient query, we first performed evaluation in a simplified simulated environment, where we assumed the correspondence with respect to a certain landmark can be established as long as the landmark is in the FoV (\ie not considering the impact of occlusion or viewpoint change).
Experiments in a more realistic environment are shown in \Sec\ref{subsec:planning}.

In the following experiment, we generated $1000$ random landmarks in a $10\si{m} \times 10\si{m} \times 5\si{m}$ area.
We further built the proposed FIF within a smaller $9\si{m} \times 9\si{m} \times 4\si{m}$ region with the size of a voxel set to $0.5\si{m}$, resulting in $\sim 16000$ voxels.
A pinhole camera model with $90 \si{deg}$ horizontal FoV was used.
With this setup, we tested the proposed FIFs using both the information factor \eqref{eq:general_info_pif} and the trace factor \eqref{eq:general_trace_pif},  with different visibility approximations, namely
\begin{itemize}
  \item Quadratic approximation that satisfies $v_\alpha=0.5, 0.8$ according to \eqref{eq:quadratic_conditions} (see \Fig\ref{fig:sim_quad_vis}).
  \item GP with $N_s = 30, 50, 70, 100, 120, 150$ trained to approximate the sigmoid visibility function \eqref{eq:sigmoid_vis_func} with $k_s = 15$.
\end{itemize}
We use the notation ``{V}-{M}-{$v_\alpha$/Ns}'' to denote a specific map representation, where {V} (\textbf{Q}uadratic or \textbf{G}P) stands for the visibility approximation, {M} (\textbf{I}nformation or \textbf{T}race) the field type, and the last number is the boundary value for quadratic approximation or the number of samples for GP approximation.
The results (the FIM and different metrics) computed from the landmarks using the exact pinhole camera model was used as the groundtruth (denoted as ``PC'').
Since the memory and time complexity of the quadratic approximation do not depend on $v_\alpha$,
we report related results for $v_\alpha = 0.5$ only and denote them as ``Q-I'' and ``Q-T'' directly.

\subsubsection{Complexity and FIM Accuracy}
To evaluate the query time and accuracy, we randomly sampled $200$ poses within the area where FIF had been constructed.
From these poses, the following were tested:
\begin{itemize}
  \item Compute the full FIM from the nearest voxel (only for FIFs using the information factors). This is to study the validity of the visibility model, excluding the impact of the voxel size and interpolation.
  \item Compute different FIM metrics using the interpolation from the nearby voxels. This is to simulate the practical use cases where the discretization resolution is limited.
\end{itemize}

\PAR{Complexity}
The query times for different settings, along with the time and memory required to build the FIFs are reported in \Tab\ref{tb:sim_complexity}.
For GP, the cases where $N_s = 50, 100, 120$ are omitted for brevity, since they follow the same trend.
In terms of query time, which is the motivation of the proposed method, all types of FIFs took much shorter time per query compared with the point cloud method.
On the other hand, the speedup at the query time comes at the cost of additional building time and memory.
In terms of different types of FIFs, quadratic approximations are faster and require less memory than GP, and trace factors are more efficient than information factors in terms of both memory and query time (last row of \Tab\ref{tb:sim_complexity}).
In addition, increasing the number of samples in GP increases the memory footprint and the query time.
Note that the query of $\det$, $\mineig$ and $\trace$  are several times more expensive than computing the FIM in our experiment, mainly due to the interpolation mentioned above (\ie accessing up to 8 surrounding voxels).

\PAR{FIM Accuracy}
To evaluate the accuracy of the computed FIM $\hat{\H}$ from the proposed information fields, we computed the relative difference with respect to the groundtruth FIM $\H$ calculated using the exact pinhole camera model from the landmarks:
\begin{equation}
  e_\text{FIM} = \lVert \hat{\H} - \H \rVert_{{F}}/{\lVert \H \rVert_{{F}}},
  \label{eq:fim_rel_diff}
\end{equation}
where $\lVert\rVert_{{F}}$ denotes the Frobenius norm. The results are reported in \Tab\ref{tb:sim_fim_acc}.
It can be seen that, for quadratic approximations, both $v_\alpha$ values tested shows rather large error.
This is not surprising, considering the long tail or negative values for large $|\theta|$ as shown in \Fig\ref{fig:sim_quad_vis}.
In terms of the GP approximations, even for GP with only $30$ samples, the recovered FIM is much more accurate than quadratic approximations.
Moreover, increasing the number of samples in GP in general decreases the difference with respect to the exact FIM.
Note that increasing the number of GP samples infinitely will \emph{not} reduce the error to zero, because our GP visibility models are designed to approximate a simplified camera model instead of the exact one.

\begin{figure}[t]
  \centering
  \includegraphics[width=0.98\linewidth]{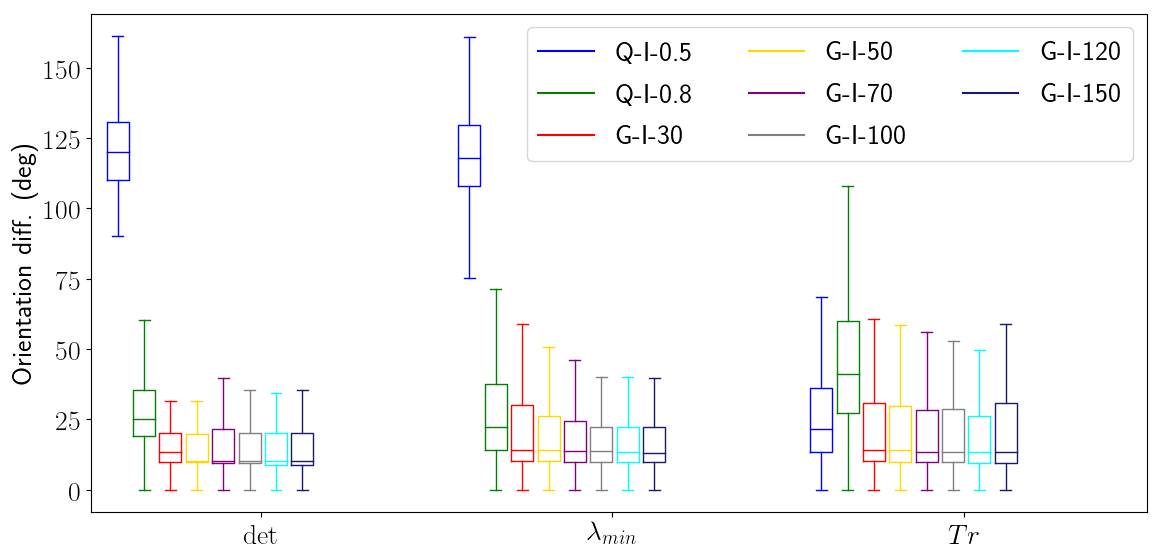}
  \caption{The differences of optimal views determined by different types of FIFs with respect to the ones determined by the point cloud. Results using determinant, the smallest eigenvalue and the trace of the FIM are shown.
  }
  \label{fig:sim_orient_boxplot}
\end{figure}

\subsubsection{Relative Measures}
In terms of motion planning, it is also of interest to check the relative values of the queried information metrics, in addition to the FIM errors mentioned above.
To this end, we performed a series of experiments about how the information metrics change with different poses.
Note that we report only the results for the information factors, since the trace factors will yield the same results as querying the trace of the FIM from the information factors.

\begin{figure}[t]
  \centering
  \begin{subfigure}[]{0.32\linewidth}
    \includegraphics[width=\linewidth, trim={2cm 2.5cm 2cm 2cm},clip]
    {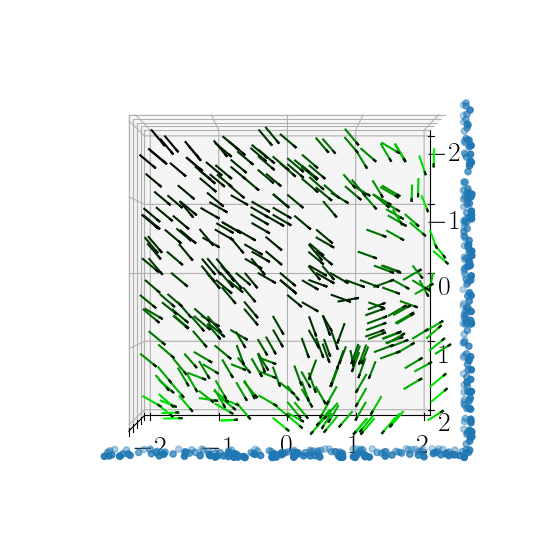}
    \caption{PC-$\det$}
  \end{subfigure}
  \begin{subfigure}[]{0.32\linewidth}
    \includegraphics[width=\linewidth, trim={2cm 2.5cm 2cm 2cm},clip]
    {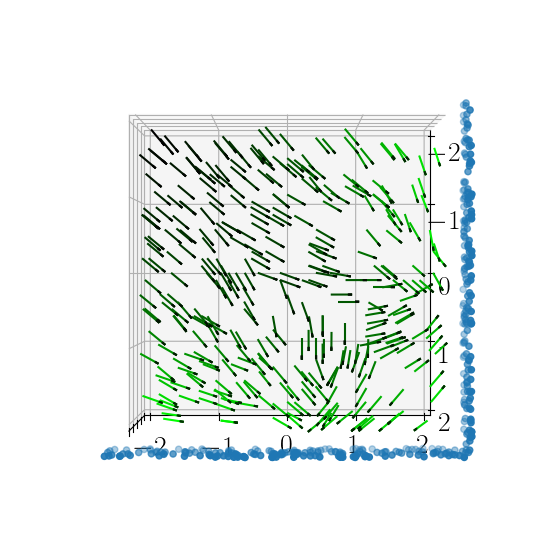}
    \caption{G-$\det$-70}
  \end{subfigure}
  \begin{subfigure}[]{0.32\linewidth}
    \includegraphics[width=\linewidth, trim={2cm 2.5cm 2cm 2cm},clip]
    {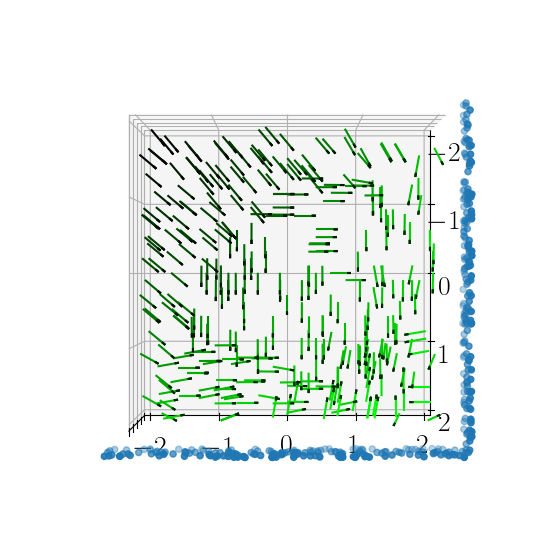}
    \caption{Q-$\det$-0.5}
  \end{subfigure}
  \begin{subfigure}[]{0.32\linewidth}
    \includegraphics[width=\linewidth, trim={2cm 2.5cm 2cm 2cm},clip]
    {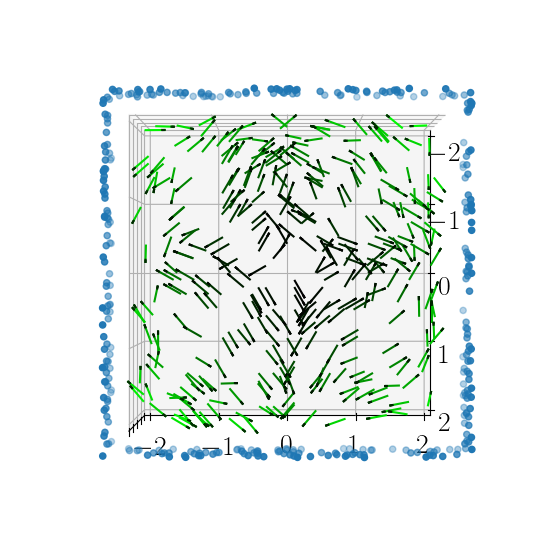}
    \caption{PC-$\det$}
  \end{subfigure}
  \begin{subfigure}[]{0.32\linewidth}
    \includegraphics[width=\linewidth, trim={2cm 2.5cm 2cm 2cm},clip]
    {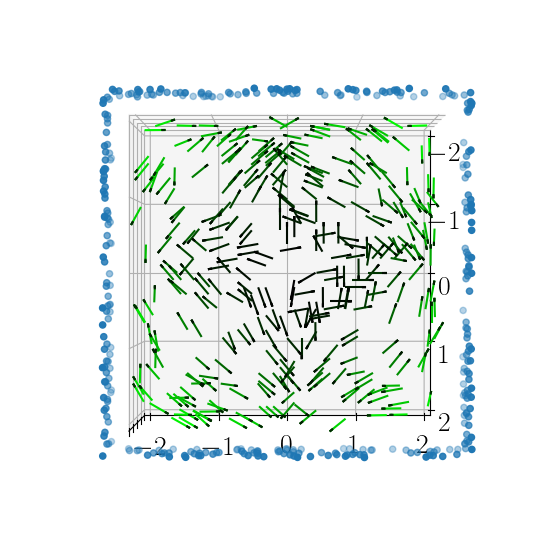}
    \caption{G-$\det$-70}
  \end{subfigure}
  \begin{subfigure}[]{0.32\linewidth}
    \includegraphics[width=\linewidth, trim={2cm 2.5cm 2cm 2cm},clip]
    {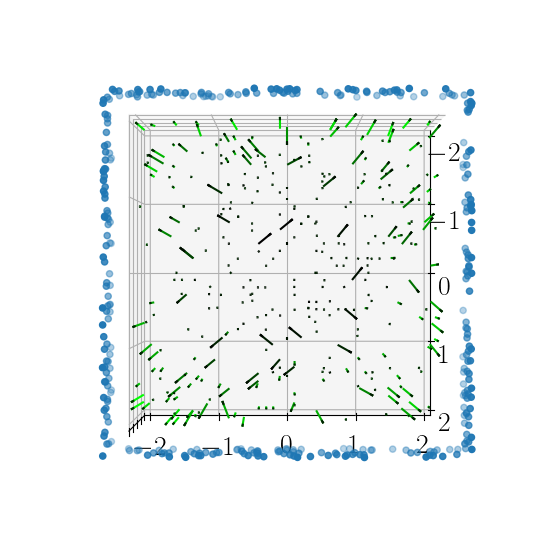}
    \caption{Q-$\det$-0.5}
  \end{subfigure}
  \begin{subfigure}[]{0.32\linewidth}
    \includegraphics[width=\linewidth, trim={2cm 2.5cm 2cm 2cm},clip]
    {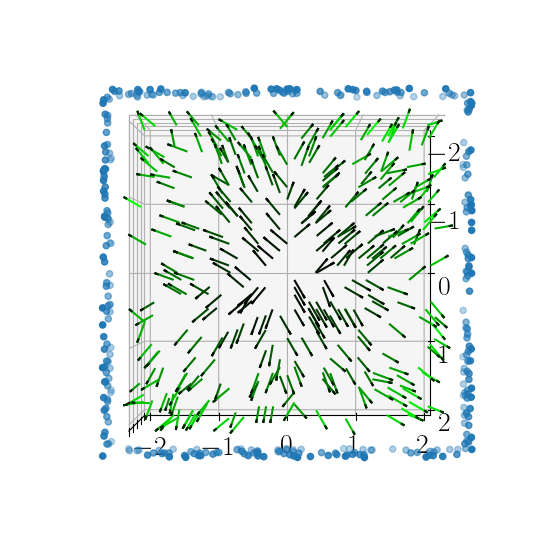}
    \caption{PC-$\trace$}
  \end{subfigure}
  \begin{subfigure}[]{0.32\linewidth}
    \includegraphics[width=\linewidth, trim={2cm 2.5cm 2cm 2cm},clip]
    {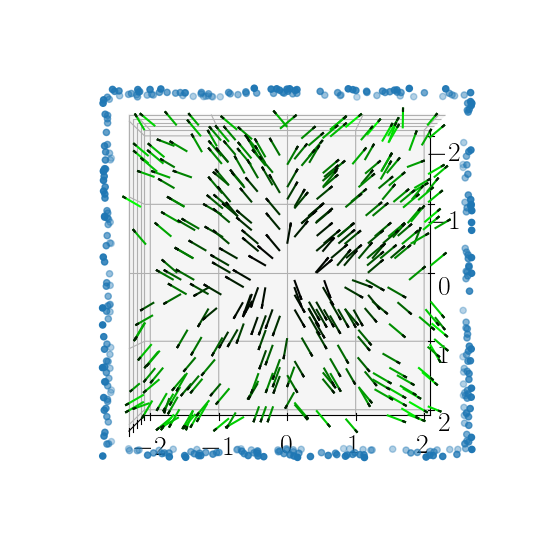}
    \caption{G-$\trace$-70}
  \end{subfigure}
  \begin{subfigure}[]{0.32\linewidth}
    \includegraphics[width=\linewidth, trim={2cm 2.5cm 2cm 2cm},clip]
    {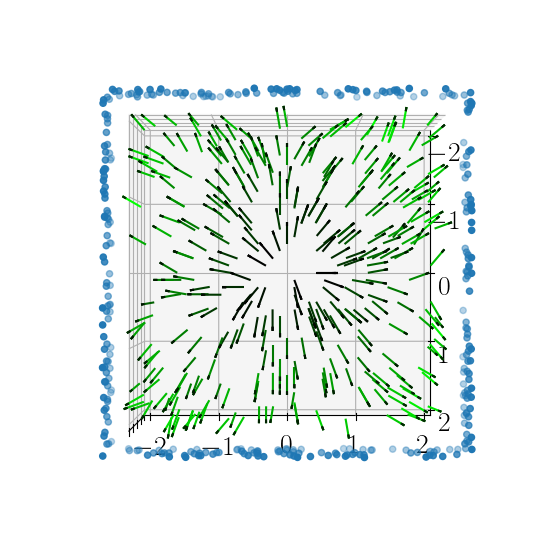}
    \caption{Q-$\trace$-0.5}
  \end{subfigure}
  \caption{
    Visualization of the information field in simulated scenes for the trace and determinant metrics.
    Blue circles are 3D landmarks, and each line segment stands for one optimal view direction.
    Brighter color means better localization quality.
    \textbf{Left}: point cloud with the exact camera model;
    \textbf{Middle}: GP approximation ($N_s=70$);
    \textbf{Right}: quadratic approximation with $v_\alpha=0.5$.
    Note the obvious failure case (f) for the combination of the quadratic model and the determinant, where the optimal views are vertical to the $xy$ plane.
    }
  \label{fig:sim_viz}
\end{figure}

\PAR{Optimal Views}
In this experiment, we computed the optimal views at $200$ sampled positions. Specifically, for each position $\twc$, we densely sampled the rotations, calculated different information metrics at the poses consisting of $\twc$ and the sampled rotations,
and determined the optimal view as the one that maximizes certain information metric.
We calculated the difference of the optimal views determined from the FIFs with respect to the ones determined by computing information metrics from the point cloud.

The differences between the optimal views calculated from FIFs and the point cloud, using different information metrics, are shown in \Fig\ref{fig:sim_orient_boxplot}.
The GP approximations are in general more accurate in determining the optimal views than the quadratic approximations.
Increasing the number of samples in GP leads to better results, but seems to saturate at around $N_s = 70$.
For quadratic approximations, the choice of $v_\alpha$ has a large impact on the optimal orientation difference.
$v_\alpha=0.8$ shows better performance than $v_\alpha=0.5$ with the determinant and the smallest eigenvalue, but using the trace with $v_\alpha=0.5$ results in smaller orientation difference.

We further computed the optimal views in simulation environments with specific landmarks layout, and several examples are shown in \Fig\ref{fig:sim_viz}.
The results are consistent with the aforementioned experiment (using randomly generated landmarks).
Intuitively, the optimal view at a position should point to the area where the landmarks are concentrated and close to the position, which is the case for the results from the point cloud and GP.
Using quadratic approximations shows larger discrepancy with respect to the point cloud, and even counter-intuitive results (\Fig\ref{fig:sim_viz} (f)).
This is probably due to the weaker expressive power of the quadratic model, as shown by the heavy tail or negative visibility for bearing vectors far away from the optical axis (values for large $|\theta |$ in \Fig\ref{fig:sim_quad_vis}).

\begin{figure}[t]
  \centering
  \begin{subfigure}[]{0.49\linewidth}
    \includegraphics[width=\linewidth]{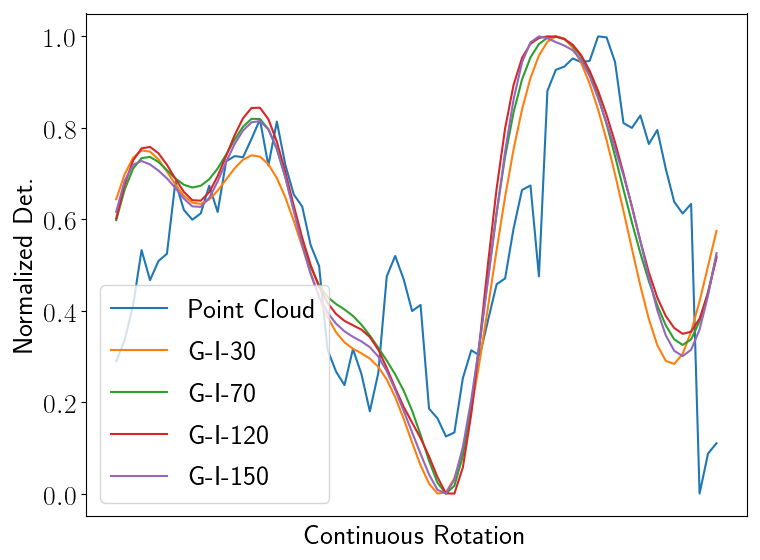}
  \end{subfigure}
  \begin{subfigure}[]{0.49\linewidth}
    \includegraphics[width=\linewidth]{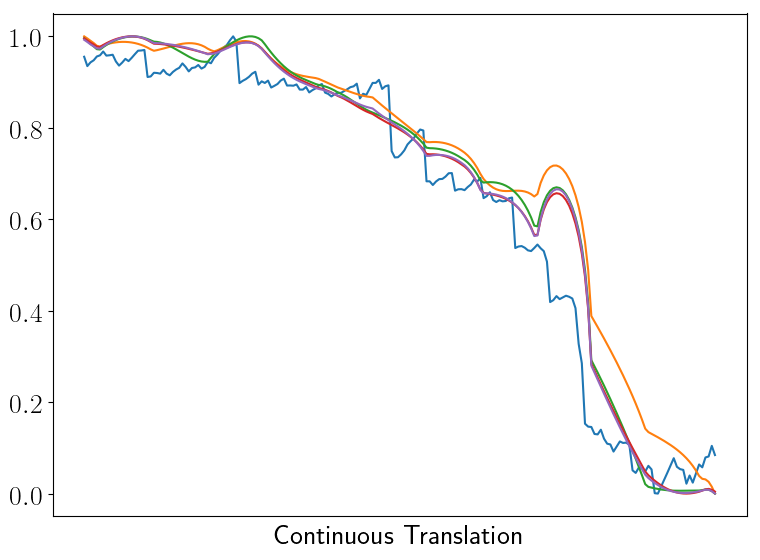}
  \end{subfigure}
  \caption{The evolution of information metric for continuous pose change.}
  \label{fig:cont_motion_info}
\end{figure}

\PAR{Smoothness}
In this experiment, we selected two continuous trajectories inside the FIF: 1) pure rotation around a fixed axis; 2) pure translation along a straight line.
We then calculated the information metrics along the two trajectories.
The evolution of the determinant (normalized to $0 - 1$ for visualization) for several FIFs and the point cloud is plotted in \Fig\ref{fig:cont_motion_info}.
Other information metrics also exhibited similar behaviors, and thus the results are omitted.
It can be seen that, while the overall trend from the FIFs are similar to the point cloud, the results from the FIFs are obviously smoother.
This property is especially important for optimization-based motion planning, as the optimization is less likely to be stuck in local minimums.
This is due to fact that the visibility approximations are differentiable, whereas the actual visibility model is not.

\begin{figure*}[t!]
  \centering
  \includegraphics[width=0.8\linewidth]{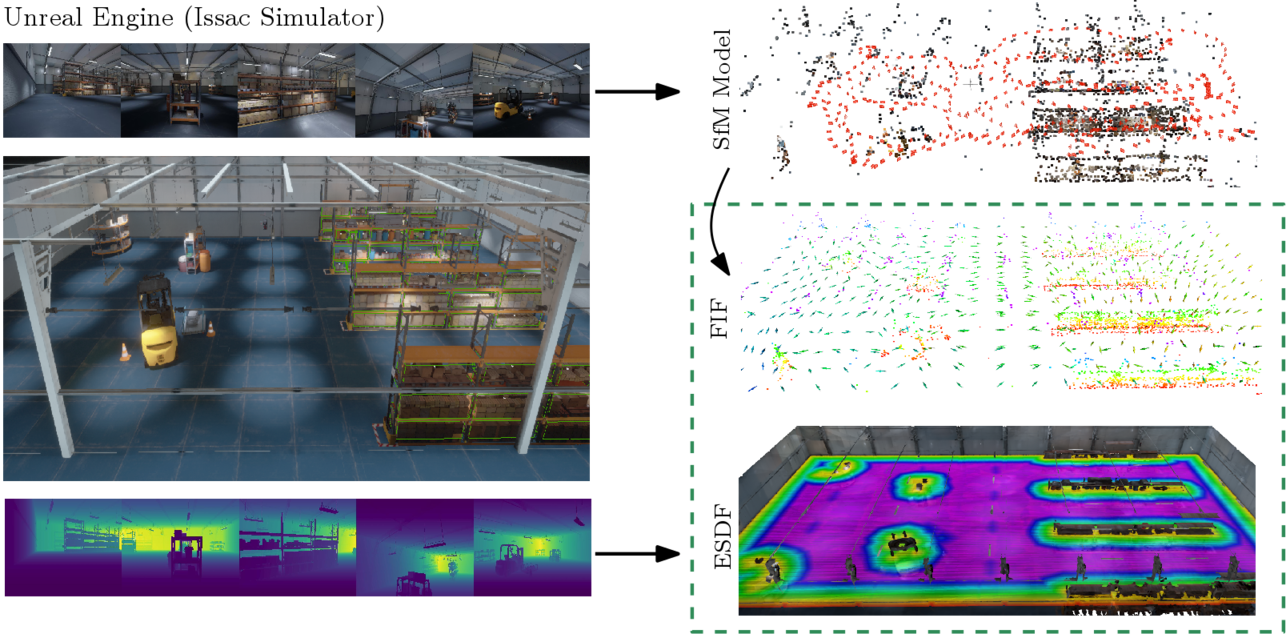}
  \caption{
    Creating different maps from the photorealistic simulation for the motion planning experiments.
    Images and depth maps were rendered from the Unreal Engine and were used, together with the camera poses, to build a SfM model (via COLMAP) and an ESDF map (via Voxblox) respectively.
    Then we built the proposed Fisher Information Field from the SfM model.
    The FIF and ESDF were then used in different perception-aware motion planning experiments.
  }
  \label{fig:exp_mapping}
\end{figure*}

\subsubsection{Summary and Discussion}
In the above experiments, we thoroughly tested different visibility approximations with both information and trace factors, which were proved to have much shorter query time than using the point cloud directly.
In general, the accuracy of the proposed FIF increases with more expensive visibility approximations (quadratic vs. GP, increasing number of GP samples).
This indicates the scalability of the proposed method: one can choose different types of visibility approximations considering the required performance and the computational resource at hand.
In addition, the trace factors proves to be significantly more efficient than the information factors, which might be of interest for computationally constrained platforms.

While GP approximations can achieve satisfactory results, the performance of the quadratic approximations, due to its limited expressive power, is not clear.
On the one hand, having a long tail or negative values for large $|\theta|$ is definitely not ideal for recovering the FIM accurately (\Tab\ref{tb:sim_fim_acc}).
On the other hand, the quadratic approximations are shown to have certain potential, if the relative values of localization quality is of interest (\eg selecting from a set of motion primitives).
This, however, seems to be scene dependent (\Fig\ref{fig:sim_viz} (c) and (f)), which is not favorable in general.

\begin{table}[t]
\scriptsize{
  \centering
  \begin{tabular}{c c c c c | c c | c c}
  \toprule
  \multicolumn{5}{c|}{FIF}&  \multicolumn{2}{c|}{ESDF} & \multicolumn{2}{c}{\# landmarks}\\
  \begin{tabular}{@{}c@{}}Voxel\\size\end{tabular} &
   \begin{tabular}{@{}c@{}}Q-I\\mem.\end{tabular}&
   \begin{tabular}{@{}c@{}}Q-T\\mem.\end{tabular}&
   \begin{tabular}{@{}c@{}}GP-I\\mem.\end{tabular}&
   \begin{tabular}{@{}c@{}}GP-T\\mem.\end{tabular}&
   \begin{tabular}{@{}c@{}}Voxel\\size\end{tabular}& Mem. & r2-a20 & r1-a30\\
  \midrule
  1.0\si{m} & 108\si{M} & 3\si{M}  & 578\si{M} & 17\si{M} &
  0.1\si{m}& 485\si{M}&
  3470 & 1445\\
  \bottomrule
  \end{tabular}
}
  \caption{
  Specifications of different maps for the photorealistic simulation environment ($\sim 50\si{m}\times30\si{m}\times9\si{m}$).
  The memory of the FIF does not change with the number of landmarks, and thus only one number is listed.
  }
  \label{tb:plan_sim_spec}
\end{table}

\subsection{Motion Planning}
\label{subsec:planning}
We further applied the proposed FIF to different motion planning algorithms in photorealistic simulation.
The experimental setup is described below.

\PAR{Photorealistic Simulation}
We used the Nvidia Issac simulator \footnote{\url{https://developer.nvidia.com/isaac-sim}} for photorealistic simulation.
We only used the rendering capability of the simulator, which is done by Unreal Engine\footnote{\url{https://www.unrealengine.com/}}.
Rendering images at desired camera poses was achieved via integrating UnrealCV \cite{Qiu17acmicm} with the simulator.
The built-in map \emph{warehouse} (see \Fig\ref{fig:exp_mapping}) was adapted and used in our experiments.
The environment is of approximately $50\si{m}\times 30\si{m} \times 9\si{m}$.

\begin{table*}[t!]
  \centering
  \begin{tabular}{ l l p{1.2cm} p{1.2cm} p{1.2cm} p{1.2cm} p{1.2cm} p{1.2cm} p{1.2cm} p{1.2cm} p{1.2cm} p{1.2cm} p{1.2cm}}
    \toprule
    & cfg.      & No Info.
    & PC-$\det$ & PC-$\trace$ & GP-$\det$ & GP-$\text{Trace}$ &
     Quad.-$\det$-0.5 & Quad.-$\trace$-0.5 &               
     Quad.-$\det$-0.8 & Quad.-$\trace$-0.8
    \\
    \midrule
    \parbox[t]{2mm}{\multirow{3}{*}{\rotatebox[origin=c]{90}{r2-a20}}} &
    Bottom  & 73\% & 0\%X & 31\% & 0\% & 17\% & 0\% & 0\% & 38\% & 31\% \\
    &
    Diagonal  & 0\% & 0\%X & 0\% & 0\% & 0\% & 0\% & 11\% & 0\% & 0\% \\
    & 
    Top & 44\% & 0\% & 29\% & 0\% & 57\% & 13\%  & 15\% & 20\% & 55\% \\
    \midrule
    \parbox[t]{2mm}{\multirow{3}{*}{\rotatebox[origin=c]{90}{r1-a30}}} &
    Bottom & 79\% & 0\%X & 25\% & 0\%X & 15\% & 0\%X  & 38\% & 7\% &36\%   \\
    &
    Diagonal  & 20\% & 0\%X & 75\% & 0\%X & 27\%  & 16\%X & 52\%  & 16\%& 18\% \\
    & 
    Top & 89\% & 0\%X & 33\% & 0\%  & 53\%  & 15\%  & 29\%  & 33\% & 69\%\\
    \bottomrule
  \end{tabular}
  \caption{
    Failure rates of localizing the rendered images on the shortest path from $\rrts$ using different types of FIFs.
    The results using two set of 3D points ``r2-a20'' and ``r1-a30'' are listed, where the former contains more points than the latter.
    ``No Info.''  denotes the case where the Fisher information was not considered in $\rrts$.
    An ``X'' denotes that $\rrts$ reported no valid solution (\eg due to the lack of landmarks in the environment).
    Note the numbers have variations across different runs of the experiment, but the comparison among different map and metric combinations stays similar.
  }
  \label{tb:rrt_fail_rate}
\end{table*}

\PAR{Planning Algorithms}
We chose two representative motion planning algorithms: $\rrts$ \cite{Karaman11ijrr} (implemented in \cite{sucan12ram}) and trajectory generation for quadrotors \footnote{\url{https://github.com/ethz-asl/mav_trajectory_generation}} \cite{burri15iros}.
$\rrts$ is a sampling-based method, whereas the trajectory generation for quadrotors relies on nonlinear optimization.
We adapted these algorithms to incorporate the information from the proposed FIF, which are described in the following sections.

\PAR{Prerequisite: Mapping the Environment}
We first mapped the environment to get different maps. In particular:
\begin{itemize}
  \item For collision avoidance, we chose to use Euclidean Signed Distance Field (ESDF) implemented in Voxblox \cite{Oleynikova17iros}.
        We densely sampled camera poses from the environment and fed the poses and depth to Voxblox to build the ESDF.
        The dense sampling is not necessary though: a more realistic exploration trajectory could also yield an ESDF that is sufficient for planning.
  \item For building the FIFs, we need sparse landmarks that can be used for localization.
        For this purpose, we manually control the camera to move around the environment to collect a series of images.
        We then fed the images and the corresponding poses to COLMAP \cite{Schoenberger16cvpr} to build a Structure from Motion (SfM) model.
        The 3D landmarks were then used to build different types of FIFs.
        To determine which landmarks are visible from a certain pose, we filtered the landmarks by the difference with respect to the average view direction in the SfM model and the depth map described below.
  \item To determine the visibility of the landmarks more accurately, we densely rendered the depth maps at the camera poses from a regular 3D grid. The depth maps were used to identify the occluded landmarks. This, in practice, could be replaced by multi-view stereo.
\end{itemize}
The mapping process and the visualization of different maps are shown in \Fig\ref{fig:exp_mapping}.
In addition, to study the impact of the number of landmarks, we further generated two SfM models and the corresponding FIFs.
The first one only contains the landmarks that has less than 2\si{px} average re-projection errors and has at least two views with larger than $20$\si{deg} parallax.
The thresholds for the second one were set to $1\si{px}$ and $30\si{deg}$.
The second SfM model contains less but, in principle, more accurate landmarks.
The two setups are denoted as ``r2-a20'' and ``r1-a30'' respectively.
The detailed specifications of the different maps used in our planning experiments are listed in \Tab\ref{tb:plan_sim_spec}.

\begin{figure}[t]
  \centering
  \includegraphics[width=0.99\linewidth]
  {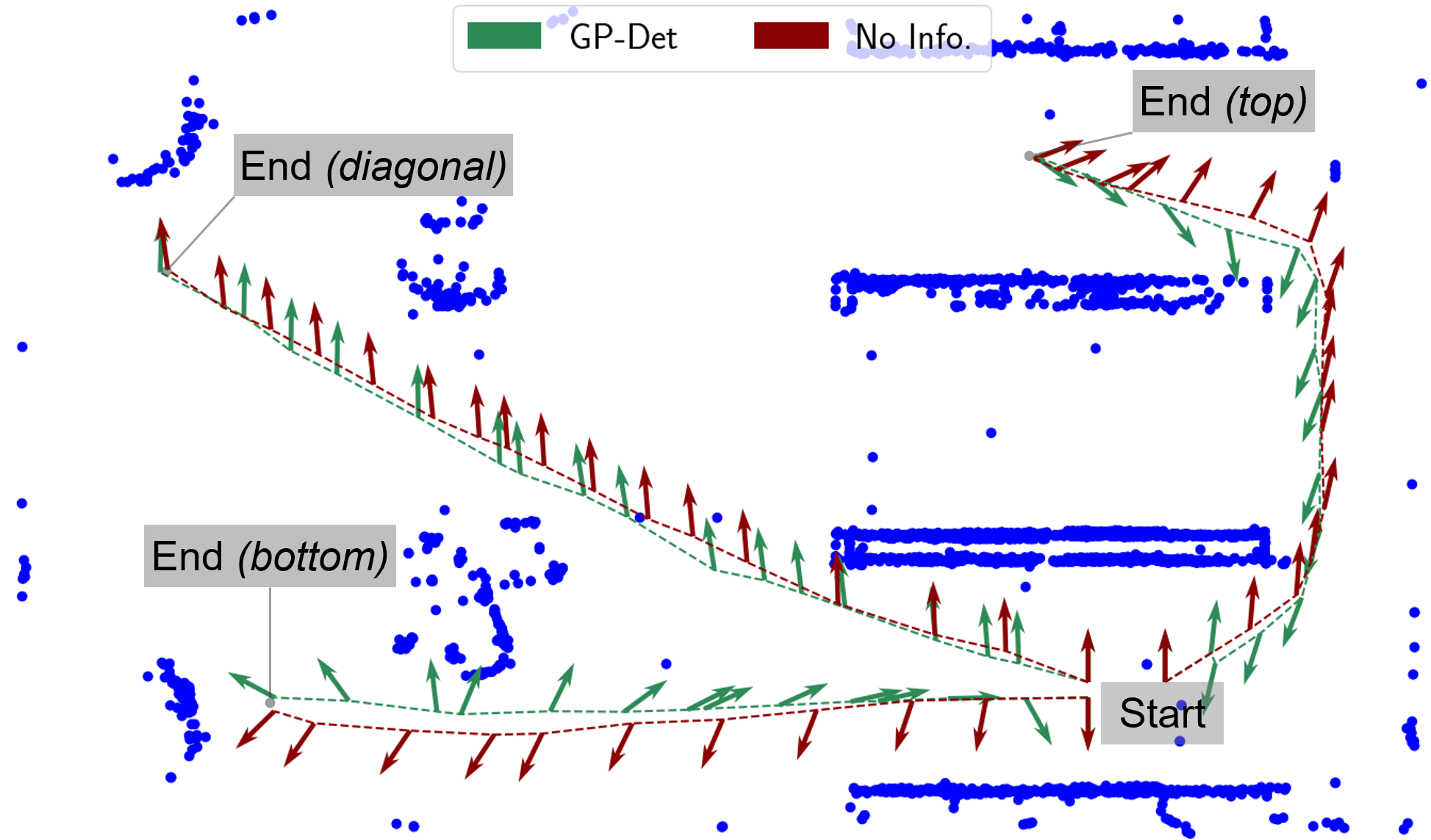}
  \caption{
    Top view of the $\rrts$ results in the simulation environment in \Fig\ref{fig:exp_mapping}  (landmark setup ``r2-a20'').
    The results of three planning configurations (\emph{bottom}, \emph{diagonal}, and \emph{top}) are shown, together with their start and end positions.
    The blue points are the landmarks, the dashed lines are the final paths from $\rrts$, and the arrows along the paths indicate the corresponding camera view directions.
  }
  \label{fig:rrt_r2_a20_top}
\end{figure}

\PAR{Workflow and Evaluation}
To test different motion planning algorithms, we followed the same workflow:
\begin{enumerate}
  \item run the motion planning algorithm
  \item sample poses from the planned motion
  \item render images at the sampled poses
  \item localize the rendered images in the SfM model using the image registration pipeline from COLMAP
\end{enumerate}
Whether the rendered images can be successfully registered and the localization accuracy are used as the evaluation metric about how much the motion planning algorithm respect the perception quality.

\PAR{Tested FIFs}
Both GP and quadratic visibility approximations were tested, using both the information and trace factors.
Since the results for the quadratic approximations are inconclusive in simulation, we tested both $v_\alpha=0.5$ and $0.8$.
In terms of the time and memory of the quadratic approximations,
we again only reported the time and memory for $v_\alpha=0.5$ for brevity.
For GP visibility approximation, we chose $N_s = 70$ based on the simulation results.
Using $N_s = 50$ produced similar results in our experiment and was more efficient.
As for the information metrics, we experimented with the determinant (calculated from the information factor) and the trace (calculated from the trace factor).
The information metrics calculated from the point cloud were used as the baseline.
The notation ``\textbf{Map}-\textbf{Metric}'' is used to label the result.

\subsubsection{$\rrts$}
The state space of $\rrts$ was set to 4 DoF: position and yaw.
It spanned the horizontally range of the \emph{warehouse} and was set to $2\si{m}$ in height.
The path lengths in the state space, in terms of position and yaw, were used as the objective to minimize, and the weights of the two costs were chosen experimentally.
Both ESDF and FIF were used to check the state validity.
In particular, the minimal distance to obstacles (from ESDF) was set to $2\si{m}$, and the information threshold (see \Sec\ref{subsec:fif_motion_planning}) was calculated by assuming that at least $10$ landmarks in $1\si{m}$ to $3\si{m}$ meter range in the FoV of the camera are needed to have a valid localization.
The planner was set to run for $500\si{s}$, regardless of whether a valid path was found.
Three planning settings (\ie start and end states) were tested, denoted as \emph{bottom}, \emph{diagonal} and \emph{top} (see \Fig\ref{fig:rrt_r2_a20_top}).

\PAR{Localization Failure Rate}
We rendered images from the poses of the vertices on the shortest path in the final tree spanned by $\rrts$ and registered the images in the SfM model.
Since we used the FIF as a state validity checker, we computed the percentages of the images that failed to be localized, shown in \Tab\ref{tb:rrt_fail_rate}.
First, the failure rates are higher in ``r1-a30'', which contains less landmarks for localization, and there are more cases where $\rrts$ failed to find a valid path as well, due to the stricter perception constraint.
Second, in general, considering Fisher information in $\rrts$ helps to reduce the failure rates, which can be seen by comparing ``No Info.'' with the other columns.
Third, ``PC-$\det$'' and ``PC-$\trace$'' both use the exact camera model, but the latter shows worse performance (in some cases even worse then ``No. Info''). This indicates that the trace of the FIM, despite its efficiency, may be a weaker indicator of the localization/pose estimation quality than the determinant. 
This is also validated by the worse performance of the trace than the determinant with both GP and quadratic visibility approximations.
Fourth, in terms of the methods that used the determinant, ``PC-$\det$'' should give the best performance (due to the use of the exact camera model), which is validated by the $0$ failure cases.
``PC-$\det$'' also has highest of number of experiments where $\rrts$ reported no solution, which indicates using ``PC-$\det$'' put a stricter criterion about whether the image from a pose can be localized.
Finally, among the combinations that used the proposed FIF and the FIM determinant, GP outperforms the quadratic approximations, and $v_\alpha=0.5$ shows better accuracy than $v_\alpha=0.8$ between the two quadratic approximations.
Notably, GP with determinant is the only FIF that has $0$ failure cases.
On the other hand, the comparison of ``GP-$\trace$'' and ``Q-$\trace$-*'' is inconclusive.
In \Fig\ref{fig:rrt_r2_a20_top} we plot the final paths of ``GP-$\det$'' and ``No Info.'' as a qualitative example.
Intuitively, with the information from the FIF, $\rrts$ prefers view directions towards area with more landmarks.

\begin{figure}[!t]
  \centering
  \includegraphics[width=\linewidth]{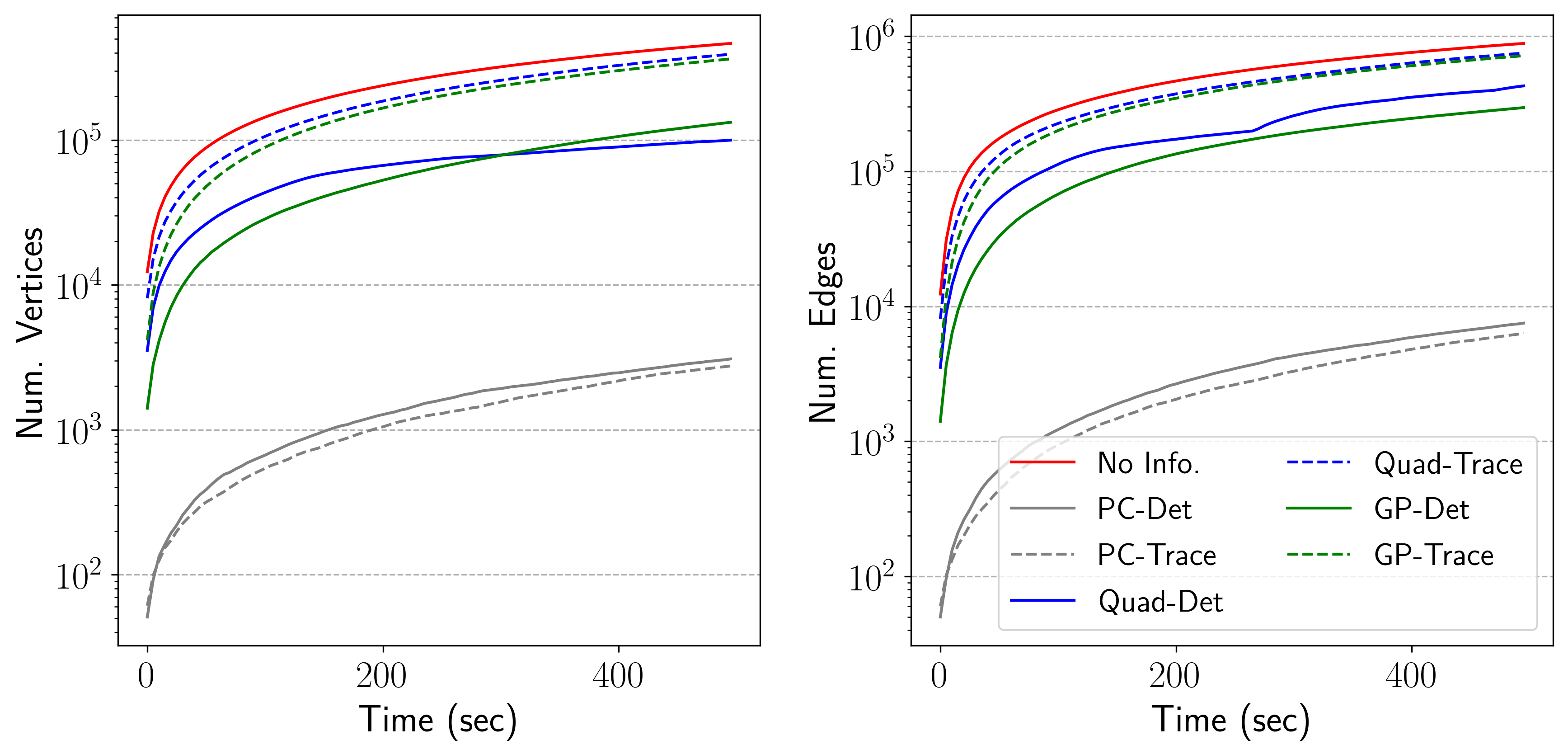}
  \caption{
    The number of vertices and edges in the $\rrts$ tree with respect the planning time.
    The plot is generated for the \emph{bottom} planning configuration in ``r2-a20''.
    Note that the $y$-axis is in $log_{10}$ scale.
  }
  \label{fig:rrt_vert_edges}
\end{figure}

\begin{figure}[!t]
  \centering
  \begin{subfigure}[]{0.49\linewidth}
    \includegraphics[width=\linewidth, trim={0cm 1cm 1.3cm 0.5cm}, clip]
    {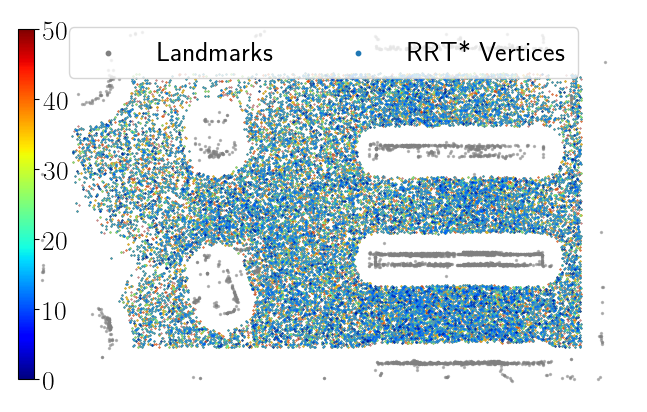}
    \caption{GP-$\det$ in ``r2-a20''}
  \end{subfigure}
  \begin{subfigure}[]{0.49\linewidth}
    \includegraphics[width=\linewidth, trim={0cm 1cm 1.3cm 0.5cm}, clip]
    {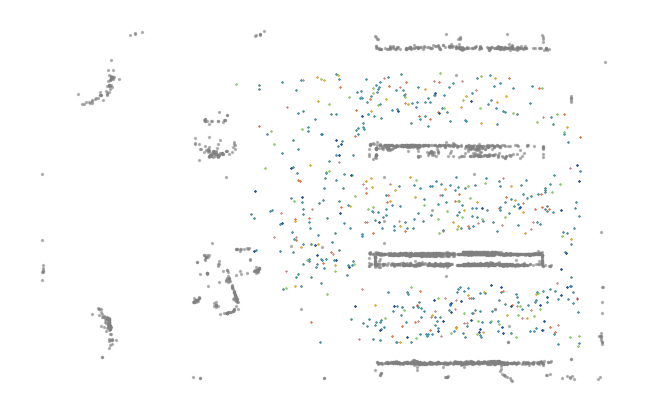}
    \caption{PC-$\det$ in ``r2-a20''}
  \end{subfigure}

  \begin{subfigure}[]{0.49\linewidth}
    \includegraphics[width=\linewidth, trim={0cm 1cm 1.3cm 0.5cm}, clip]
    {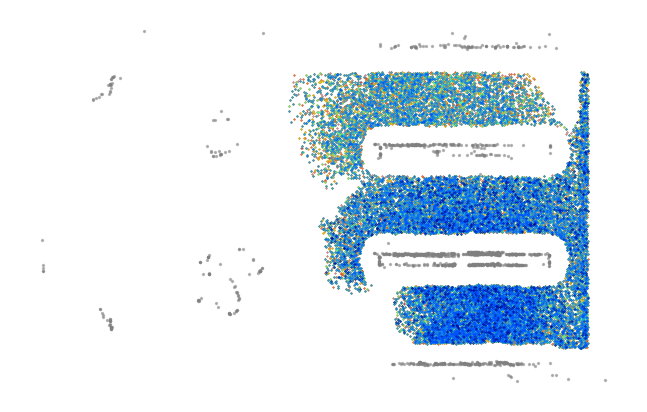}
    \caption{GP-$\det$ in ``r1-a30''}
  \end{subfigure}
  \begin{subfigure}[]{0.49\linewidth}
    \includegraphics[width=\linewidth, trim={0cm 1cm 1.3cm 0.5cm}, clip]
    {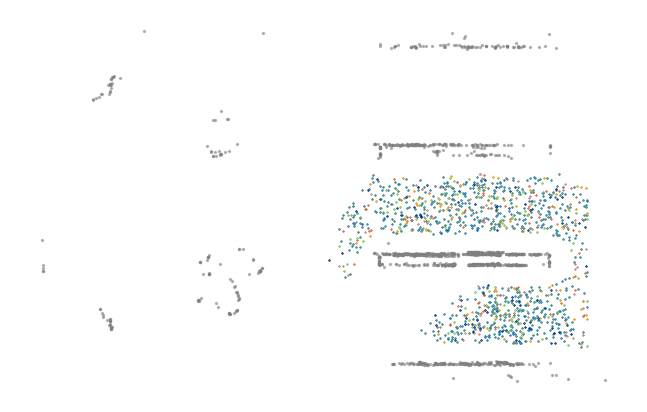}
    \caption{PC-$\det$ in ``r1-a30''}
  \end{subfigure}
  \caption{
    The $\rrts$ vertices that were explored for the first 50 seconds.
    The color of a $\rrts$ vertices indicate when the vertices were added to the tree (indicated by the colorbar).
    The gray points denote the landmarks in the environment.
  }
  \label{fig:rrt_vert_top}
\end{figure}

\PAR{Efficiency}
As for the efficiency of different map representations, we plot the number of vertices and edges in the tree spanned by $\rrts$ with respect to the time spent.
Since the results of different planning configurations are similar, we show one example in \Fig\ref{fig:rrt_vert_edges}.
All types of FIFs tested are at least one order of magnitude faster in terms of the number of vertices that can be explored for the same time.
In addition, the quadratic model is more efficient than GP, and computing the trace from the FIFs is faster than computing the determinant.
Qualitatively, in \Fig\ref{fig:rrt_vert_top}, we plot the vertices for the first $50$ seconds for the \emph{bottom} planning configuration in both ``r2-a20'' and ``r1-a30'' together with the landmarks.
Comparing the left column (``GP-$\det$'') and the right column (``PC-$\det$''), the vertices explored using the proposed FIF cover a larger area with a higher sampling density than using the point cloud.
Comparing the first row with the second, we can see that decreasing the number of landmarks effectively reduce the region where the poses are considered to be able to be localized.
This is also potentially useful to identify the ``perception traps'' in a given environment.

\subsubsection{Trajectory Optimization}
\begin{figure*}[t!]
  \centering
  \begin{subfigure}[]{0.248\linewidth}
    \includegraphics[width=\linewidth, trim={0cm 0cm 1.5cm 0cm}, clip]
    {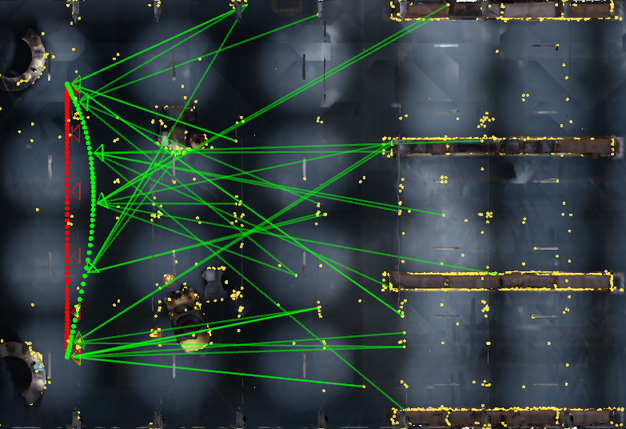}
    \caption{side}
  \end{subfigure} 
  \hspace{-0.5em}
  \begin{subfigure}[]{0.248\linewidth}
    \includegraphics[width=\linewidth, trim={0cm 0cm 1.5cm 0cm}, clip]
    {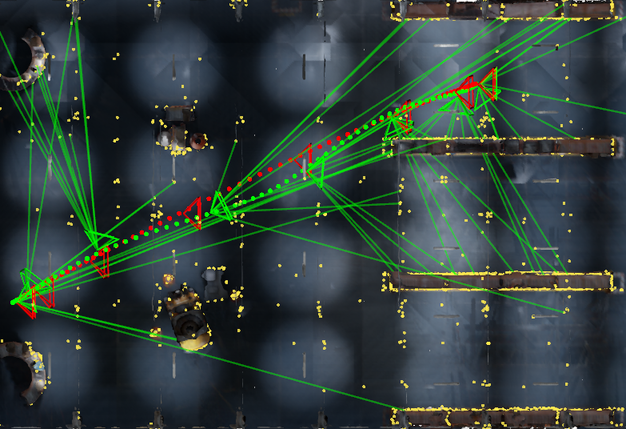}
    \caption{top}
  \end{subfigure}
  \hspace{-0.5em}
  \begin{subfigure}[]{0.248\linewidth}
    \includegraphics[width=\linewidth, trim={0cm 0cm 1.5cm 0cm}, clip]
    {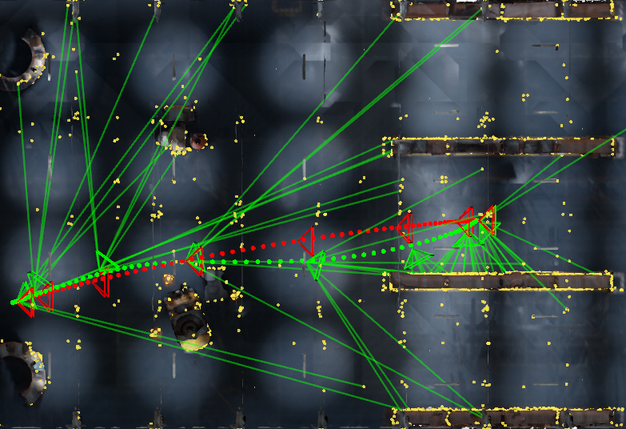}
    \caption{middle}
  \end{subfigure}
  \hspace{-0.5em}
  \begin{subfigure}[]{0.248\linewidth}
    \includegraphics[width=\linewidth, trim={0cm 0cm 1.5cm 0cm}, clip]
    {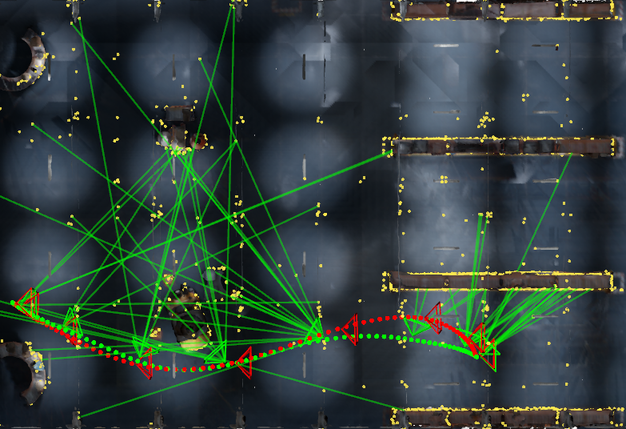}
    \caption{bottom}\label{fig:traj_opt_viz_bottom}
  \end{subfigure}
  \caption{
    The comparison of the optimized trajectories using the proposed Fisher Information Field (\textcolor{green}{green})
     and without considering the Fisher information (\textcolor{red}{red}).
    The poses sampled at a constant time interval are visualized as points of the corresponding color.
    The yellow points are the landmarks for localization, and the lines denote the potentially matchable landmarks considered in the trajectory optimization.
    GP visibility approximation and the determinant of the FIM are used.
    All trajectories start from the left part of the scenario.
    Note that only the top views are shown, but the trajectories are optimized in the 3D space.
    For example, in \emph{side}, the green trajectory is higher than the red one, favoring the landmarks located on the ceiling.
  }
  \label{fig:traj_opt_viz}
\end{figure*}

Following the standard practice \cite{Mellinger11icra}, we used a piecewise 4 DoF polynomial (5 segments) to represent a continuous-time trajectory for quadrotors.
Given start and end states, we first initialized the trajectory using \cite{Richter13isrr} and used it as an initial value for further nonlinear optimization.
In the nonlinear optimization, we considered the position and yaw dynamic cost of the quadrotors, the collision potential cost as in \cite{Oleynikova16iros, Oleynikova18ral}, and the information potential cost from the FIFs.
Specifically, the dynamic costs were calculated in closed-form, derived from the polynomial coefficients, and the collision potential cost and the information potential cost were calculated as the integral along the trajectory, with a sampling interval of $0.1\si{sec}$.
For the information potential cost \eqref{eq:info_pot_cost}, $200$ landmarks within $0.3\si{m}$ to $1.0\si{m}$ were considered sufficient and used to calculate $\epsilon_{\fim}$ for different information metrics and map representations.
The weights among these costs were chosen experimentally but kept fixed for all the experiments.
The optimization was modeled as a general unconstrained optimization problem using Ceres\footnote{\url{http://ceres-solver.org/}}, and the default optimizer parameters were used.
For each trajectory optimization, we let the optimizer run for maximum $100$ iterations.
Similar to the $\rrts$ experiment, we chose four sets of start and end states, namely \emph{top}, \emph{middle}, \emph{bottom} and \emph{side} (see \Fig\ref{fig:traj_opt_viz}).
The duration of the trajectory was set to $10$\si{sec}.

\begin{figure}[t]
  \centering
  \begin{subfigure}[]{0.99\linewidth}
    \includegraphics[width=\linewidth]{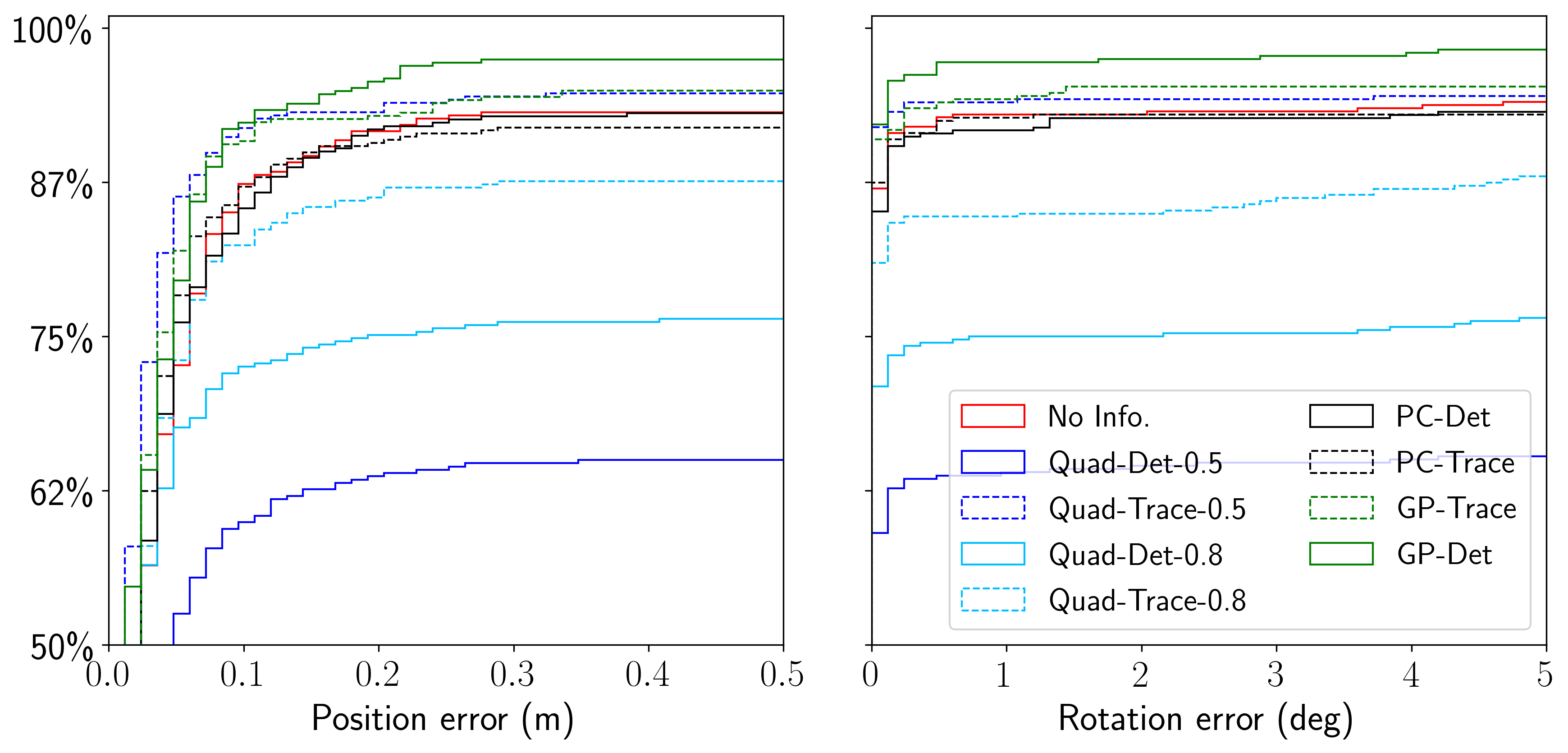}
    \caption{r2-a20}
  \end{subfigure}
  \begin{subfigure}[]{0.99\linewidth}
    \includegraphics[width=\linewidth]{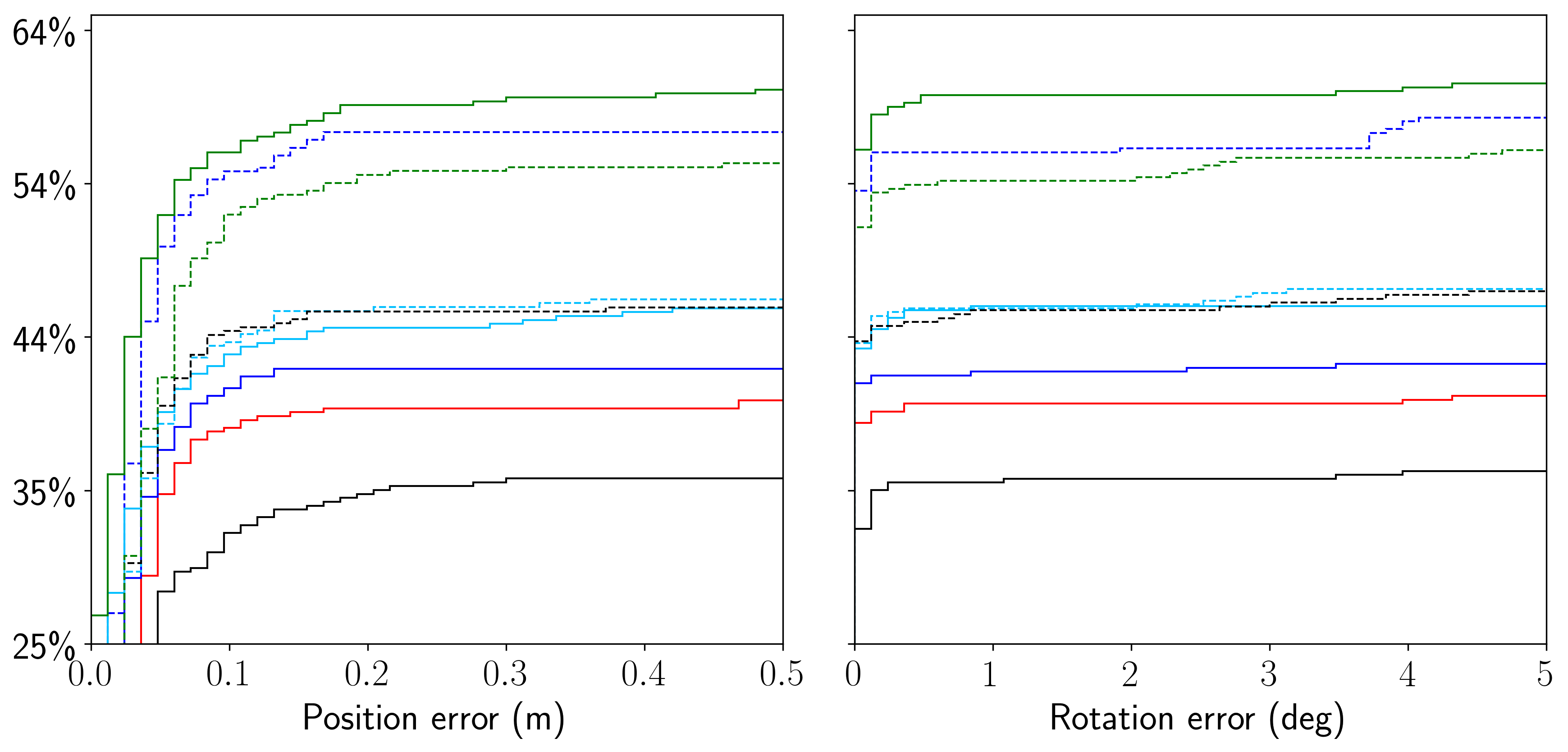}
    \caption{r1-a30}
  \end{subfigure}
  \caption{
    Cumulative histograms of the localization error of the images rendered from the optimized trajectories using different map representations in ``r2-a20'' and ``r1-a30''.
    Each point $(X, Y)$ on the curve denotes there are $X\%$ images that were able to be localized below $Y\si{m}$ (or $Y\si{deg}$) error.
  }
  \label{fig:traj_opt_acc}
\end{figure}

\PAR{Localization Accuracy}
After the optimizer converged or reached the maximum number of iterations, we sampled poses from the trajectory by $0.1\si{sec}$ time interval and rendered images from these poses.
Since the FIM was used as an optimization objective, we calculated the localization error with respect to the true poses for evaluation.
The cumulative histograms of the position and rotation errors, aggregated over all planning settings, are shown in \Fig\ref{fig:traj_opt_acc}.
Similar to the $\rrts$ experiment, we observed that decreasing the number of landmarks reduced the localization accuracy, and the benefit of considering the Fisher information becomes more significant, as shown by the larger margin.
Compared with ``No Info.'', considering the Fisher information in general improves the localization accuracy, with a few exceptions.
In particular, the quadratic approximations with the determinant of  the FIM performed significantly worse in ``r2-a20'' and slightly better than  ``No Info.'' in ``r1-a30''.
The accuracy of ``Quad-$\trace$-$0.8$'' also shows similar behavior.
Overall, ``GP-$\det$'', ``GP-$\trace$'' and ``Q-$\trace$-0.5'' are the best performing combinations, where the GP approximation with the determinant of the FIM consistently shows the highest localization accuracy.
This is consistent with results of the optimal view experiment in \Fig\ref{fig:sim_orient_boxplot}.
Interestingly, computing information metrics from the point cloud overall shows no obvious improvement with respect to ``No Info.''.
Notably, ``PC-$\det$'', which was the best performing combination in the $\rrts$ experiments, achieved lower accuracy than not considering the Fisher information at all.
We further observed that the optimization using the point cloud tended to terminate prematurely (see \Tab\ref{tb:traj_aver_iter_times}), which is possibly due to the discontinuity shown in \Fig\ref{fig:cont_motion_info} caused by the exact camera model.
Examples of the optimized trajectories (from ``GP-$\det$'') are shown in \Fig\ref{fig:traj_opt_viz}.
Intuitively, including the Fisher information in the trajectory optimization force the camera to orient towards and move closer to areas with more landmarks (\eg the shelves), resulting in more matchable landmarks in the camera FoV and higher localization accuracy.

\begin{table}[t]
  \centering
  \begin{tabular}{ l c c c c }
    \toprule
    & \multicolumn{2}{c}{r2-a20} & \multicolumn{2}{c}{r1-a30} \\
    & $\overline{\text{iter.}}$ & $\overline{\text{time}}$ (s)  &
    $\overline{\text{iter.}}$ & $\overline{\text{time}}$ (s) \\
    \midrule
    No Info.    & 59.0  & 0.057 & 59.0 & 0.054 \\
    \midrule
    PC-$\det$   & 12.8  & 44.49 & 9.0  & 12.68 \\
    PC-$\trace$ & 17.5  & 65.74 & 16.3 & 31.10 \\
    GP-$\det$    & 100.0 & 1.35  & 62.5 & 0.99 \\
    GP-$\trace$  & 99.0  & 0.39  & 93.8 & 0.40 \\
    Quadratic-$\det$    & 85.8  & 0.35  & 61.8 & 0.26 \\
    Quadratic-$\trace$  & 90.8  & 0.13  & 79.0 & 0.12 \\
    \bottomrule
  \end{tabular}
  \caption{
  Average number of iterations and optimization time over all planning settings in the trajectory generation experiment.
  The maximum number of iterations was set to $100$.
  }
  \label{tb:traj_aver_iter_times}
\end{table}

\PAR{Efficiency}
The average number of iterations and the optimization time are listed in \Tab\ref{tb:traj_aver_iter_times}.
Among the methods that consider the Fisher information, calculating information metrics from the point cloud takes the most time with the lowest number of iterations, which indicates that the evaluation of the information metrics using the point cloud is far more expensive than the proposed map representations.
Moreover, we suspect that the lower number of iterations indicates the the optimization terminated prematurely, considering the lower localization accuracy shown in \Fig\ref{fig:traj_opt_acc}.
Similar to the results in the $\rrts$ experiment, we observe that using GP is more time-consuming than the quadratic model.
The trace factors, despite the slightly worse localization accuracy, is very efficient: with the quadratic model,  it is only around two times slower than the case where no Fisher information was considered.

\subsubsection{Summary and discussion}
Different from the simplified simulation in \Sec\ref{subsec:sim}, we applied the proposed FIF to different motion planning algorithms in a realistic simulation.
The images were rendered using a photorealistic game engine, and the localization accuracy was evaluated using mainstream visual localization techniques.
It can be seen that, in general, integrating FIF helps improve the localization quality, in terms of the successful rate and localization accuracy.
Compared with the standard practice of using the point cloud, our method is at least one order-of-magnitude faster, and the differentiable/smooth visibility approximations additionally bring better performance in the trajectory optimization experiment.
In terms of different visibility approximations, similar to the simplified simulation in \Sec\ref{subsec:sim}, the GP approximation shows satisfactory performance, and the performance of the quadratic approximations highly depends on $v_\alpha$ and the layout of the landmarks.
While it is not conclusive that which $v_\alpha$ and metric should be used in general, the quadratic approximation is shown to be useful in certain cases (\eg ``Quad-$\trace$-0.5'' in the trajectory optimization experiment).
Next, we further discuss several aspects of our method.

\PAR{Generalizability}
{First}, we would like to highlight that the proposed FIF is not specific to certain motion planning algorithms.
In particular, we chose two representative motion planning algorithms (\ie sampling-based and optimization-based).
Moreover, off-the-shelf open source implementations of these algorithms were used through their existing interface, without specific customization for our map representation.
{Second}, it is relatively easy to build a specific map with ``perception traps'' to show the benefit of perception-aware motion planning algorithms compared with standard ones (as discussed in our previous work \cite{Zhang18icra}).
However, in the above experiment, we tried to avoid artificial corner cases to evaluate our method in a relative realistic setup.
The improved performance indicates the proposed method is a useful tool in general.

\PAR{Offline Mapping}
The process of constructing the FIFs in this section is a relatively complicated process, since it requires the knowledge of the scene depth as well as the average view direction for each landmark to determine accurately whether a landmark can be matched from a given pose.
While it certainly constitutes a barrier for building the FIFs incrementally online, this also justifies our proposal of having a dedicated map for localization/perception quality: since quantifying the localization quality from the point cloud is an expensive process, a dedicated map that can be built offline (where the efficiency is less important) and used for efficient planning online would be useful.
Besides, in many practical applications, the robot operates in a known environment, and building a map offline is thus a reasonable choice.

\PAR{Planning with Multiple Objectives}
As shown in our experiments, planning tasks usually need to consider multiple (possibly conflicting) objectives altogether, which is a non-trivial task.
The planning setup used in the experiment is probably not optimal in terms of both the objectives (\eg looking sideways in \Fig\ref{fig:traj_opt_viz_bottom} is not good for detecting unknown obstacles ahead) and the way of combining them (\eg a simple weighted sum was used in trajectory optimization).
Our goal, however, is to show that our map is efficient, differentiable and suitable for different planning algorithms, which we believe is thoroughly validated in our experiments.
If necessary, our map representation can be readily used with additional objectives and alternative methods of combining different objectives (\eg multiobjective search in \cite{Ichter17isrr}).
\subsection{Incremental Update}
\label{subsec:inc_update}
As mentioned in \Sec\ref{subsec:fif_map}, due to the additive natural of the FIM, the information and trace factors can be potentially updated as new landmarks are added/deleted from the environment.
It is, however, limited by the fact that it is difficult to accurately determine whether the correspondence with respect to a landmark can be established from a certain position in an online fashion.
Nevertheless, for relatively simple environments, incrementally building the FIF can still give reasonable result.

To illustrate this, we ran a stereo VIO pipeline that consists of an efficient frontend \cite{Forster17troSVO} and an optimization backend \cite{Leutenegger15ijrr} on the images and inertial measurements collected on a drone \cite{Burri15ijrr} and built the proposed FIF incrementally from the output of the VIO.
In particular, we added voxels to the FIF around the estimated pose of the drone whenever the it moved to the regions that were not already covered by the FIF.
Meanwhile, when new landmarks were estimated by the VIO, we identified the voxels that could observe the landmarks by a simple distance check and updated the positional factors in these voxels incrementally.
This thus simulated an exploration scenario, where the FIF is built incrementally as the robot explores an unknown environment relying on only on-board sensing.
We refer the reader to the accompanying video for the corresponding qualitative results.

%% file: sections/conclusion.tex
\section{Conclusion and Future Work}
\label{sec:conclusion}
In this work, we proposed the first dedicated map representation, the Fisher information field, for considering localization accuracy in perception-aware motion planning.
For a known environment, the proposed map representation pre-computes the rotation-invariant component of the Fisher information and stores it in a voxel grid.
At planning time, the Fisher information matrix (and related metrics) can be computed in constant time, regardless of the number of landmarks in the environment.
We validated the effectiveness and advantages of the FIF by applying it to different motion planning algorithms, namely $\rrts$ and trajectory optimization.
Integrating the proposed map in motion planning algorithms was shown to increase the localization success rate and accuracy.
All the variants of the proposed map showed $1\sim2$ order-of-magnitude shorter planning time than the point cloud.
In trajectory optimization, the proposed map representation, in addition of being far more efficient, achieved better localization accuracy than the point cloud, thanks to the fact that our map is differentiable.

The pre-computation, which is the key for the efficiency of the proposed map, is possible due to the special form of the visibility model \eqref{eq:general_vis_approx} we enforced.
Following this form, polynomial and GP approximations were explored in the paper.
In particular, the quadratic polynomial model \eqref{eq:quad_vis_approx} and GP model \eqref{eq:gp_vis_approx} with different number of samples were implemented and tested.
Different information metrics from the FIM were also tested.
Using the combination of GP ($50\sim70$ samples) with the FIM determinant showed overall the best performance among all the variants.
The advantage of the quadratic approximation, due to its limited expressive power, is less clear in comparison.
However, the efficiency (in both memory and query time) of the quadratic approximation can be still appealing in certain situations.
For example, in the trajectory optimization experiment, the combination of the quadratic model ($v_\alpha=0.5$) and the FIM trace was only slightly worse than GP with the FIM determinant but $\sim10$ times more efficient (\Tab\ref{tb:traj_aver_iter_times}).

Along the same line of the discussion about different visibility approximations, exploring other different visibility approximations that satisfy \eqref{eq:general_vis_approx} or more general \eqref{eq:info_separation} would be one interesting future direction.
For example, the form of the quadratic approximation \eqref{eq:quad_vis_approx} could be naturally extended to higher order polynomials, which could be further fitted to a smooth visibility function such as \eqref{eq:sigmoid_vis_func}.
Higher order polynomials involve more terms (\ie more expensive) but have better expressive power, which could possibly fall between the quadratic approximation and the non-parametric approximation in terms of the complexity and performance tradeoff.

One key step of building the proposed map from the point cloud is to accurately predict whether a landmark can be matched in images (and used for visual localization) from a certain viewpoint.
This is, however, not an easy task in many non-trivial environments, which requires much information in addition to the positions of the landmarks.
To overcome this limitation, there are several interesting research directions.
First, since such process ideally requires geometric, semantic (\eg whether a landmark is on a static object and can be used for localization) and texture information, developing more advanced map representations that combine different types of information would be useful.
Such map representations would also be useful for other robotic tasks, \eg \cite{Rosinol20rss,verdoja2020arxiv}.
Second, being able to determine the matchability of landmarks without knowing the full information of the environment (or being able to update such information as the environment get better explored) would greatly extend the application scenario of the proposed method.
For example, it would allow to incrementally build an accurate FIF during the exploration of an unknown environment using the output of visual-inertial odometry as an input, as demonstrated in \Sec\ref{subsec:inc_update} for simple scene layouts.

Lastly, the proposed map representation can be viewed as a mapping from a 6 DoF pose to its localization quality.
In essence, our method implements the mapping by a combination of the interpolation over position (the voxel grid data structure) and the regression of the landmark visibility from rotation to achieve efficient evaluation of the mapping.
From a more general perspective, one could design the full mapping directly and seek suitable data structures for desired properties (efficiency and/or accuracy).
There are a variety of possible choices and reserach oppurtunities.
For example, recent advance in neural scene representations shows great potential of implicit scene representation for novel view synthesis \cite{sitzmann2019deepvoxels,sitzmann2019srns, Mildenhall20arxiv}, which could be potentially used for the purpose of modelling the localization quality as well.